\pdfoutput=1
\documentclass[11pt]{article}
\usepackage[preprint]{acl}
\usepackage{times}
\usepackage{latexsym}
\usepackage[T1]{fontenc}
\usepackage[utf8]{inputenc}
\usepackage{microtype}
\usepackage{graphicx}
\usepackage{booktabs}
\usepackage{pifont}
\usepackage{xurl}
\usepackage{amsmath}
\usepackage{amssymb}
\usepackage{tikz}
\usepackage{pgfplots}
\usepackage{colortbl}
\DeclareSymbolFont{textdigits}{T1}{ptm}{m}{n}
\DeclareMathSymbol{0}{\mathalpha}{textdigits}{`0}
\DeclareMathSymbol{1}{\mathalpha}{textdigits}{`1}
\DeclareMathSymbol{2}{\mathalpha}{textdigits}{`2}
\DeclareMathSymbol{3}{\mathalpha}{textdigits}{`3}
\DeclareMathSymbol{4}{\mathalpha}{textdigits}{`4}
\DeclareMathSymbol{5}{\mathalpha}{textdigits}{`5}
\DeclareMathSymbol{6}{\mathalpha}{textdigits}{`6}
\DeclareMathSymbol{7}{\mathalpha}{textdigits}{`7}
\DeclareMathSymbol{8}{\mathalpha}{textdigits}{`8}
\DeclareMathSymbol{9}{\mathalpha}{textdigits}{`9}
\DeclareMathSymbol{.}{\mathord}{textdigits}{`.}
\definecolor{oursrow}{gray}{0.92}
\newcommand{\grouprow}[2]{\rowcolor{oursrow}\multicolumn{#1}{l}{\emph{#2}} \\}
\usepgfplotslibrary{groupplots}
\usetikzlibrary{arrows.meta,calc}
\pgfplotsset{compat=1.17}
\makeatletter\def\Inter@scale{0.85}\makeatother
\newcommand{\figfont}{\fontfamily{Inter-LF}\selectfont}
\definecolor{vizblue}{HTML}{2A78D6}
\definecolor{vizorange}{HTML}{EB6834}
\definecolor{vizink}{HTML}{52514E}
\definecolor{vizred}{HTML}{D1362F}
\pgfplotsset{vizaxis/.style={axis lines*=left, axis line style={black!35, line width=0.4pt, -},
  major tick length=0pt, minor tick style={draw=none}, ymajorgrids, grid style={black!8, line width=0.3pt},
  tick label style={font=\fontsize{6.5}{7}\selectfont, text=vizink}, label style={font=\fontsize{7}{8}\selectfont, text=vizink},
  title style={font=\fontsize{7}{8}\selectfont, text=black!85, yshift=-2pt}, enlargelimits=false, clip=false,
  legend style={font=\fontsize{6.5}{7}\selectfont, text=vizink, draw=none, fill=none},
  every axis plot/.append style={line width=0.8pt}},
  vizline/.style={line width=0.8pt, mark=*, mark size=1.15pt}}
\pgfplotsset{every axis/.append style={
  xticklabel={\pgfmathprintnumber[assume math mode=true]{\tick}},
  yticklabel={\pgfmathprintnumber[assume math mode=true]{\tick}}}}

\title{ReListen: Training-Free Evidence Localization in Hour-Long Audio\\ with Frozen Audio Language Models}

\author{%
  Ming-Hao Hsu$^{1,\dagger}$ \quad
  Shujie Liu$^{3,*}$ \quad
  Yuxuan Hu$^{2}$ \quad
  Yuxiang Wang$^{1}$ \\
  \bfseries Yingda Shen$^{1}$ \quad
  Jinyu Li$^{2}$ \quad
  Zhizheng Wu$^{1,*}$
}

\begin{document}
\maketitle
\begingroup
\renewcommand{\thefootnote}{}
\footnotetext{$^{1}$School of Data Science, The Chinese University of Hong Kong, Shenzhen, China. $^{2}$Microsoft CoreAI, Redmond, WA, USA. $^{3}$Microsoft Research Asia, Hong Kong, China. $^*$Correspondence to: Shujie Liu $<$\texttt{shujliu@microsoft.com}$>$ and Zhizheng Wu $<$\texttt{wuzhizheng@cuhk.edu.cn}$>$. $^\dagger$Work done during an internship at Microsoft.}
\endgroup
\begin{abstract}
Audio language models are meant to answer questions from the audio itself, but an hour-long recording rarely fits in their context, and the answer often depends on a few seconds of it.
Transcribing the recording first makes it fit but discards sounds and voices, while systems that return to the audio must be trained for this behavior.
We propose ReListen, a training-free method in which a frozen audio language model finds its own evidence before answering.
In a Jev-style scan, the model scores each 20-second window by the probability that it helps answer the question, and it then answers from the highest-scoring minute.
This minute overlaps the annotated evidence far more often than transcript retrieval, for 88.5\% of VoiceGiraffe questions.
Without any change, the same scan also grounds events on LAT-Bench as well as a model trained for temporal grounding.
Answering from it matches or exceeds hearing the whole recording with a small fraction of the audio, and on VoiceGiraffe it exceeds a transcript of the whole recording by 9.2 points.
The selected evidence is also worth hearing rather than reading.
Given the same excerpts, reading their transcript matches listening on speech content, but listening is clearly more accurate on questions about sounds and voices.
Moreover, ReListen answers faster than every transcript-based system, and its memory stays constant even for recordings too long to hear whole on one GPU.
Therefore, a frozen audio language model can search hour-long recordings by itself, without a transcript or additional training.
\end{abstract}

\section{Introduction}

Audio language models answer open questions about speech and sound in clips of seconds to a few minutes \citep{chu2024qwen2audio,tang2024salmonn,gong2024ltu,ghosh2025audioflamingo3,xu2025qwen3omni}.
However, meetings, lectures, broadcasts and podcasts last an hour or more.
Benchmarks built from such recordings report that accuracy falls as duration grows \citep{ahia2025blab,he2025audiomarathon,luo2026chronosaudio,huang2026longaudiospan,ye2026voicegiraffe}.
Indeed, many models cannot take an hour of audio as input.
Those that can must still find a few relevant seconds.

\begin{figure}[t]
\centering
\begingroup\figfont%
\definecolor{ovink}{RGB}{44,62,92}%
\definecolor{ovacc}{RGB}{206,84,8}%
\definecolor{ovwave}{gray}{0.66}%
\definecolor{ovmute}{gray}{0.40}%
\newcommand{\ovf}{\fontsize{7.6}{8.6}\selectfont}%
\newcommand{\ovs}{\fontsize{6.6}{7.6}\selectfont}%
\pgfplotsset{ovaxis/.style={anchor=south west, scale only axis, width=7.04cm, clip=false, enlargelimits=false}}%
\begin{tikzpicture}[x=1cm, y=1cm, line cap=round,
  >={Stealth[length=3.4pt, width=3.2pt]},
  badge/.style={circle, fill=black!80, text=white, inner sep=0pt, minimum size=8.6pt,
    font=\fontsize{6.2}{6.2}\selectfont\bfseries},
  head/.style={anchor=base west, inner sep=0pt, font=\ovf\bfseries},
  note/.style={inner sep=0pt, font=\ovs, text=ovmute},
  reader/.style={draw=black!72, line width=0.45pt, rounded corners=1.6pt, fill=white,
    inner xsep=3.6pt, inner ysep=2.4pt, font=\ovf}]
\useasboundingbox (0,-0.10) rectangle (7.66,6.50);

\node[anchor=north west, inner sep=0pt, text width=7.66cm, align=left, font=\footnotesize] at (0,6.50)
  {\textbf{Question.}\enspace\textit{How many distinct speakers are heard during the pineapple-pizza and cold-pizza debate?}};

\fill[black!7] (2.787,5.13) -- (3.204,5.13) -- (3.204,3.50) -- (7.66,2.90) -- (0.62,2.90) -- (2.787,3.50) -- cycle;

\node[badge] at (0.15,5.585) {1};
\node[head] at (0.37,5.50) {Jev-style scan};
\node[reader, anchor=base west] (r1) at (2.08,5.50) {frozen model};
\draw[->, black!72, line width=0.45pt] (r1.east) -- ++(0.30,0);
\node[note, anchor=base west] at ([xshift=0.38cm]r1.base east) {log-odds of ``yes'' per window};

\foreach \t/\l in {0/0, 20/20, 40/40, 60/60, 80/80} {
  \draw[black!55, line width=0.3pt] ({0.62+7.04*\t/84.46},5.13) -- ++(0,0.045);
  \node[anchor=south, inner sep=0pt, font=\ovs, text=ovmute] at ({0.62+7.04*\t/84.46},5.20) {\l};
}
\begin{axis}[ovaxis, hide axis, at={(0.62cm,4.69cm)}, height=0.44cm, xmin=0, xmax=84.46, ymin=-1.05, ymax=1.05]
\addplot[draw=none, fill=ovwave] table[x=minute, y=amp] {figures/overview_wave.dat} \closedcycle;
\addplot[draw=none, fill=ovwave] table[x=minute, y expr=-\thisrow{amp}] {figures/overview_wave.dat} \closedcycle;
\end{axis}
\node[note, anchor=south east] at (0.50,5.20) {min};
\node[note, anchor=east] at (0.50,4.91) {audio};

\begin{axis}[ovaxis, at={(0.62cm,3.50cm)}, height=1.12cm, xmin=0, xmax=84.46, ymin=-3.7, ymax=3.9,
  axis x line=none, axis y line=left, y axis line style={-, black!55, line width=0.3pt},
  ytick={-2,0,2}, tick style={black!55, line width=0.3pt}, major tick length=1.6pt, tick align=outside,
  tick label style={font=\ovs, text=ovmute, inner sep=1.2pt}]
\draw[black!30, line width=0.3pt, densely dotted] (axis cs:0,0) -- (axis cs:84.46,0);
\addplot[ovink, line width=0.42pt, line join=round] table[x=minute, y=score] {figures/overview_scan.dat};
\addplot[only marks, mark=*, mark size=1.05pt, ovacc] coordinates {(27.667,2.80) (28.167,3.10) (28.667,2.82)};
\end{axis}
\node[note, anchor=north west] at (0.70,4.60) {score $\ell_i$};

\node[badge] at (0.15,3.215) {2};
\node[head] at (0.37,3.13) {Select};
\node[note, anchor=base east] at (7.66,3.13) {the three best windows};

\foreach \t/\l in {60/27{:}00, 120/28{:}00, 180/29{:}00, 240/30{:}00} {
  \draw[black!55, line width=0.3pt] ({0.62+7.04*\t/300},2.90) -- ++(0,-0.045);
  \node[anchor=north, inner sep=0pt, font=\ovs, text=ovmute] at ({0.62+7.04*\t/300},2.83) {\l};
}
\foreach \a in {90,120,150} {
  \fill[ovacc!16] ({0.62+7.04*\a/300},1.13) rectangle ({0.62+7.04*(\a+20)/300},2.56);
}
\begin{axis}[ovaxis, hide axis, at={(0.62cm,1.62cm)}, height=0.94cm, xmin=0, xmax=300, ymin=-2.5, ymax=4.6]
\draw[black!40, line width=0.3pt] (axis cs:0,0) -- (axis cs:300,0);
\addplot[ycomb, ovink!75, line width=0.5pt, mark=*, mark size=0.85pt]
  table[x=second, y=score, restrict expr to domain={\thisrow{selected}}{0:0}] {figures/overview_zoom.dat};
\addplot[ycomb, ovacc, line width=0.8pt, mark=*, mark size=1.15pt]
  table[x=second, y=score, restrict expr to domain={\thisrow{selected}}{1:1}] {figures/overview_zoom.dat};
\end{axis}
\begin{axis}[ovaxis, hide axis, at={(0.62cm,1.13cm)}, height=0.42cm, xmin=0, xmax=300, ymin=-1.05, ymax=1.05]
\addplot[draw=none, fill=ovwave] table[x=second, y=amp] {figures/overview_zoomwave.dat} \closedcycle;
\addplot[draw=none, fill=ovwave] table[x=second, y expr=-\thisrow{amp}] {figures/overview_zoomwave.dat} \closedcycle;
\begin{scope}
\clip (axis cs:90,-1.05) rectangle (axis cs:110,1.05) (axis cs:120,-1.05) rectangle (axis cs:140,1.05)
  (axis cs:150,-1.05) rectangle (axis cs:170,1.05);
\addplot[draw=none, fill=ovacc] table[x=second, y=amp] {figures/overview_zoomwave.dat} \closedcycle;
\addplot[draw=none, fill=ovacc] table[x=second, y expr=-\thisrow{amp}] {figures/overview_zoomwave.dat} \closedcycle;
\end{scope}
\end{axis}
\node[note, anchor=east] at (0.50,1.951) {score};
\node[note, anchor=east] at (0.50,1.34) {audio};
\draw[black!85, line width=0.8pt, line cap=butt] ({0.62+7.04*97/300},2.50) -- ({0.62+7.04*162/300},2.50);
\foreach \e in {97,162} {\draw[black!85, line width=0.5pt] ({0.62+7.04*\e/300},2.455) -- ++(0,0.09);}
\node[note, anchor=west, text=black!75] at ({0.62+7.04*162/300+0.08},2.50) {annotated evidence};

\foreach \a/\l in {90/27{:}30, 120/28{:}00, 150/28{:}30} {
  \node[anchor=north, inner sep=0pt, font=\ovs, text=ovacc] at ({0.62+7.04*(\a+10)/300},1.07) {\l};
}
\draw[ovacc, line width=0.45pt] ({0.62+7.04*90/300},0.80) -- ++(0,-0.07) -- ({0.62+7.04*170/300},0.73) -- ++(0,0.07);
\node[reader] (r3) at ({0.62+7.04*130/300},0.26) {frozen model};
\draw[->, ovacc, line width=0.5pt] ({0.62+7.04*130/300},0.73) -- (r3.north);
\node[badge] at (0.15,0.345) {3};
\node[head] at (0.37,0.26) {Listen again};
\node[note, anchor=base west] at (0.37,-0.03) {60 s, with time labels};
\node[anchor=base west, inner sep=0pt, font=\ovf] (ans) at ([xshift=0.48cm]r3.base east) {answer: \textbf{5} $\checkmark$};
\draw[->, black!72, line width=0.45pt] (r3.east) -- ++(0.36,0);
\node[note, anchor=base west] at ([yshift=-8.3pt]ans.base west) {uniform windows: 4 $\times$};
\end{tikzpicture}%
\endgroup
\caption{\textbf{ReListen on a test question about an 84-minute recording.}
All scores and answers come from the actual run.
The frozen audio language model scores each 20-second window by the log-odds of a yes when asked whether the window helps to answer the question.
The three best windows overlap the annotated evidence, and the answer from them is correct.
From three evenly spaced windows, the same model answers 4, which is wrong.}
\label{fig:overview}
\end{figure}
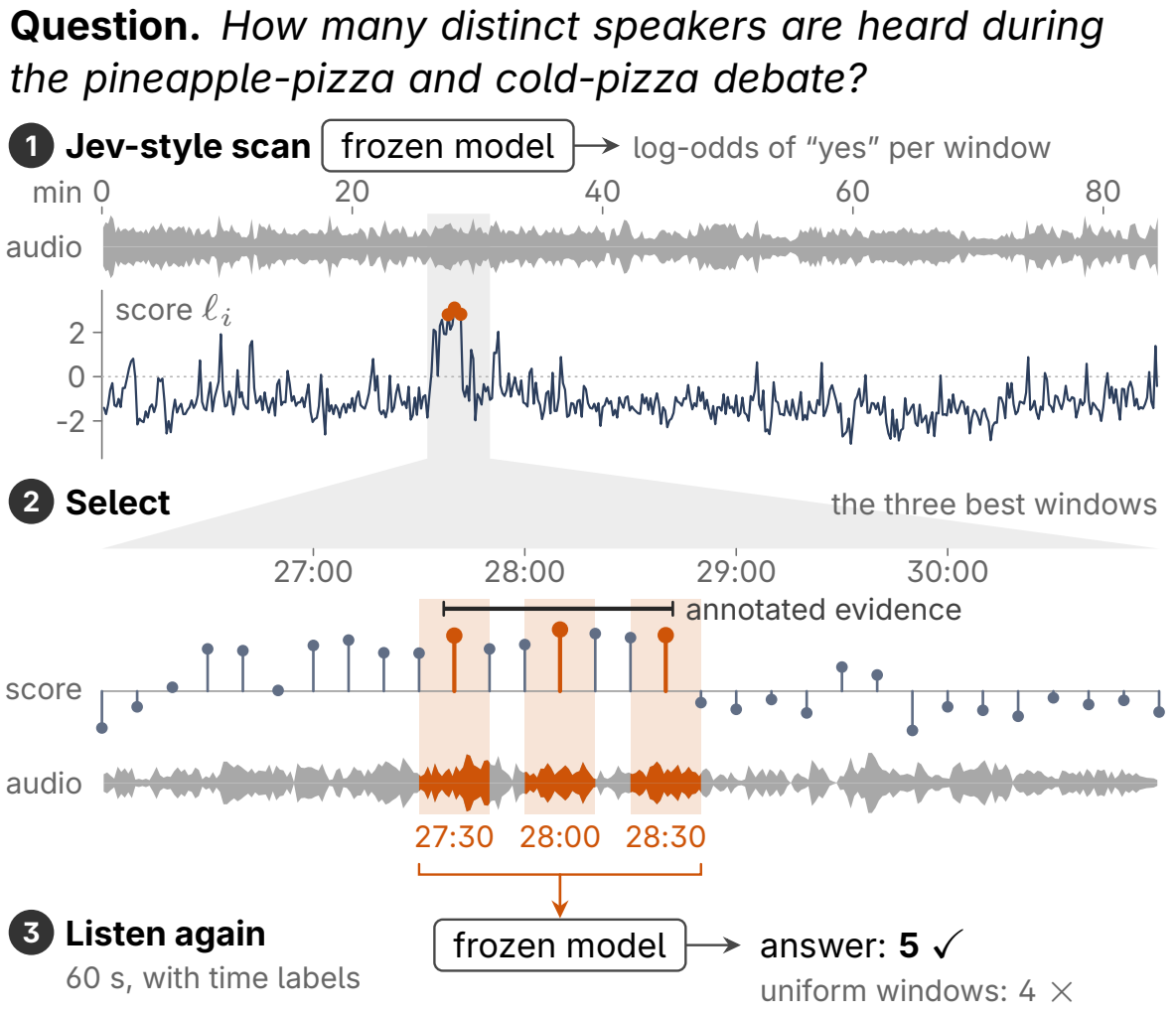

A common solution places a second system in front of the audio model \citep{chen2025wavrag,min2025speechrag,someki2026planragaudio}.
This system is either a speech recognizer whose transcript is searched or read, or a retriever over audio embeddings.
A transcript is particularly effective, since it compresses an hour of speech into text that fits the model's context.
However, the audio model then answers as a text reader.
The transcript also discards sounds, voices and timing, which many questions ask about and which audio language models are built to hear.
Other systems therefore keep the audio and train the model to return to selected parts of it.
They refine a global timeline with audio crops \citep{shao2026listeningwithtime} or request segments during reasoning \citep{wu2026echo}.
These systems must be fine-tuned for this behavior, and the training must be repeated for each new audio model.
Therefore, a useful solution should search the audio itself, but without a separate front end or additional training.

In this paper, we show that a frozen audio language model can perform this search itself.
Our key idea is to reduce the search to many small decisions that the model can already make, namely whether a short excerpt helps to answer the question.
We therefore propose ReListen, which uses one frozen audio language model both to search and to answer (Figure~\ref{fig:overview}).
ReListen splits the recording into short windows and scores each window by the model's confidence that it helps to answer the question.
The same model then listens again to the highest-scoring minute and answers.
This scoring is a Jev-style scan \citep{typesafe2026jev}: each window receives a yes/no decision read as a probability rather than generated text.
The scan examines the whole recording, while the answer uses only one minute of audio.
No component is trained.

We select the configuration on a development half of LongAudioSpan \citep{huang2026longaudiospan}.
We then evaluate it unchanged on the test half, VoiceGiraffe \citep{ye2026voicegiraffe} and LAT-Bench \citep{shao2026listeningwithtime}.
First, the scan finds evidence.
The selected minute overlaps the annotated evidence for 88.5\% of VoiceGiraffe questions, compared with 58.8\% for transcript retrieval.
On LAT-Bench, its mean temporal IoU is comparable to that of the trained LAT-Audio model (47.7 vs.\ 46.3).
Second, the selected evidence improves answers.
At the same one-minute budget, ReListen exceeds uniformly sampled audio by 8.5 and 21.3 points on LongAudioSpan and VoiceGiraffe.
On VoiceGiraffe, it is the most accurate audio input and exceeds the reader's transcript of the whole recording by 9.2 points.
On LongAudioSpan, it is on par with hearing the whole recording.
On VoiceGiraffe sound questions, listening to the selected excerpts also exceeds reading their transcript by 8.6 points.
Third, ReListen is efficient, since it needs no training, no transcript and only one minute of audio for the answer.
On recordings of about an hour, it answers a question seven to eight times faster than the reader's transcript pipeline.
Its memory also stays constant with duration, and its independent windows divide across GPUs.
Taken together, a frozen audio language model can find its own evidence in hour-long recordings, and answering from this evidence is both accurate and efficient.

\section{Related Work}
\label{sec:related}

\begin{figure*}[t]
\centering
\input{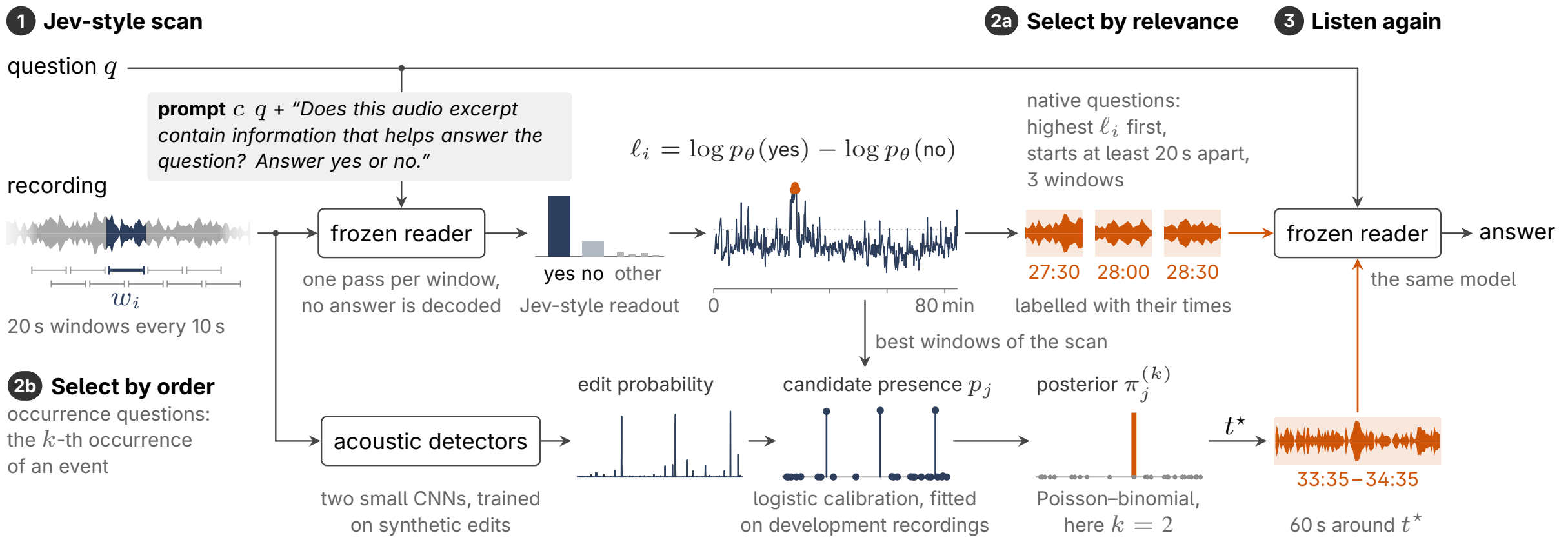}
\caption{\textbf{Architecture of ReListen.}
One frozen reader scores every window in a Jev-style scan (1).
It reads each yes/no decision from one forward pass as a probability, and it then answers from the selected audio (3).
The native-question path selects by relevance without training (2a).
The supervised occurrence extension uses trained acoustic detectors and a posterior over calibrated candidates (2b).}
\label{fig:architecture}
\end{figure*}

Answering questions about long recordings requires deciding which part of the audio to process.

\paragraph{Long-Audio Understanding.}
Prior systems fit long recordings into an audio language model in three ways.
They prune audio tokens \citep{guo2025fastlongspeech,lin2025speechprune}, retrieve over an index of audio, transcripts or events \citep{chen2025wavrag,min2025speechrag,someki2026planragaudio}, or read text descriptions of short windows \citep{huang2026longaudiospan,ye2026voicegiraffe}.
These approaches make hour-long inputs tractable and are effective when the question concerns what is said.
However, their selection relies on a separate front end, such as a recognizer, an embedding model or a captioner, which decides what the answering model receives.
In contrast, we let the answering model decide what to hear.

\paragraph{Answer Probabilities as Scores.}
Reading the probability of a yes as a relevance score is common in ranking and evaluation.
Text rerankers score query-document pairs this way \citep{nogueira2020monot5}, VQAScore scores image-text alignment \citep{lin2024vqascore}, and recent decision models such as Jev \citep{typesafe2026jev} answer typed questions with probabilities instead of text.
Closest to our scan is the localization probe of video question answering \citep{yu2023sevila}, which asks the same kind of yes/no question about video frames.
Its localizer, however, is a separately trained module on clips of 12 to 76 seconds.
In contrast, our Jev-style scan is the frozen answering model itself, applied to recordings in which one minute is at most a tenth of the audio.
Yes/no probes of audio language models also hallucinate presence \citep{kuan2024objecthallucination} and under-report recurring events \citep{dai2026tagbench}.
Therefore, for occurrence questions, we further calibrate these scores \citep{guo2017calibration}.

\paragraph{Re-Listening Audio Models.}
Recent audio models show that returning to the audio improves temporal understanding and reasoning.
LAT-Audio \citep{shao2026listeningwithtime} first constructs a global timeline and then refines it with audio crops.
Echo \citep{wu2026echo} trains Qwen2.5-Omni to request audio segments by generated timestamps during its reasoning.
It is trained by supervised fine-tuning on clips of about ten seconds, followed by reinforcement learning.
Speaker-Reasoner \citep{lin2026speakerreasoner} learns multi-turn temporal interaction for timestamped, speaker-attributed transcription and extends it with a speaker-aware cache.
These systems learn when and where to listen, but this behavior must be trained for each model, and the times they use are generated by the model.
In contrast, ReListen trains nothing and needs neither a transcript nor generated timestamps.
The times of its excerpts are simply the positions of the windows that its scan selects.
We compare with these systems in Section~\ref{sec:results} and Appendix~\ref{app:prior-art}.

\section{ReListen}
\label{sec:method}

Given a long recording and a question about it, the task is to answer the question.
For a grounding query, the task is instead to predict the interval that the query describes.
We call the frozen audio language model that answers the \emph{reader}.
\emph{Native questions} concern the natural content of a recording.
\emph{Occurrence questions} instead ask about the $k$-th occurrence of an event, such as a sound planted in the recording.
Existing systems let a separate front end decide what the reader hears.
In contrast, ReListen lets the frozen reader make this decision itself (Figure~\ref{fig:architecture}).
The reader scores every short window of the recording by whether the window helps to answer the question.
It then answers from the highest-scoring windows.
The Jev-style scan and native-question answering use no training.
We also study a supervised occurrence extension with acoustic detectors and calibration on annotated development recordings (Section~\ref{sec:occurrence}).

\subsection{Jev-Style Scan by the Reader}
\label{sec:scan}

We first cut the recording into windows $w_1, \dots, w_n$ of length $\tau$, with a hop $\delta < \tau$ between consecutive windows ($\tau{=}20$ and $\delta{=}10$ seconds in our experiments).
For each window $w_i$, the reader then receives the audio and a yes/no prompt $c$ built from the question (Appendix~\ref{app:prompts}).
For a native question, the prompt states the question, with its answer options when there are any, and asks whether the excerpt contains information that helps answer it.
For a question about an event $e$, it instead asks whether the excerpt contains $e$.
We score each window without generating a yes/no sentence.
Using the next-token distribution $p_\theta$ at the first answer position, we write this score as
\begin{equation}
\ell_i = \log p_\theta(\text{yes} \mid w_i, c) - \log p_\theta(\text{no} \mid w_i, c),
\end{equation}
where each probability sums over the forms of the answer word in each prompt language.
A positive score means that yes is more likely than no.
Higher scores rank the windows for selection.
We call this a \emph{Jev-style scan}.
Like the Jev decision model \citep{typesafe2026jev}, it answers a typed yes/no question with a probability instead of text.
All windows of a recording share the prompt and the length, so they form batches without padding.
Each window costs one forward pass.
The cost of the scan therefore grows linearly with duration at constant memory.
The window passes are independent, so they can also run in parallel on several GPUs.

\subsection{Answering Native Questions}
\label{sec:content}

For a native question, ReListen takes the windows in decreasing order of $\ell_i$.
It keeps a window only if it does not overlap any window already kept.
It stops after $b$ windows, that is, $b\tau$ seconds of audio ($b{=}3$, or 60 seconds, in our experiments).
A larger budget suits questions whose evidence spreads across the recording (Section~\ref{sec:global-results}).
The reader then answers the original question once from these excerpts in chronological order.
Adjacent excerpts are merged, and each is labelled with its start and end time in the recording, since many options name times.
Answering one question thus costs $n$ window passes and one answer from $b\tau$ seconds of audio, without any transcription.

\paragraph{Event Intervals.}
For a grounding query, ReListen outputs one interval instead of excerpts.
This \emph{expanded region} starts from the highest-scoring window and extends it left and right through consecutive windows with positive scores.
Its duration is therefore set by the scores, not by the budget $b\tau$.

\subsection{Extension to Occurrence Questions}
\label{sec:occurrence}

However, selection by relevance has an inherent limit.
A question about the $k$-th occurrence of an event is answered correctly only if every earlier occurrence is found and no spurious one is counted before it.
Yet all occurrences match the event's description equally well.
For these questions, we therefore study a supervised extension.
Some event types, such as changes of speed and volume, are not described by what the audio contains.
On development recordings, the scan's scores indeed carry little information about them (Appendix~\ref{app:occurrence}).
Therefore, two small acoustic detectors, trained on synthetic edits of the benchmark's types, propose candidate event times.
The highest-scoring windows of the scan add further candidates.
A logistic model fitted on annotated development recordings then assigns each candidate $j$ a probability $p_j$ of being a real event.
A natural rule keeps the candidates above a threshold and takes the $k$-th in time.
However, a single false alarm or miss before the target then shifts the answer by one occurrence.
Instead, the extension estimates for each candidate the probability that it is real and that exactly $k-1$ earlier candidates are real.
With the candidates sorted by time and treated as independent, this probability is
\begin{equation}
\pi_j^{(k)} = p_j \cdot \Pr\Big[\textstyle\sum_{i<j} z_i = k-1\Big],
\end{equation}
with $z_i \sim \mathrm{Bernoulli}(p_i)$.
The second factor is a Poisson-binomial probability \citep{hong2013poissonbinomial}, computed for all $j$ by dynamic programming.
The extension then selects the candidate with the largest $\pi_j^{(k)}$.
The reader then answers the remaining questions of the chain from one minute of audio around this candidate (Appendix~\ref{app:occurrence}).

\section{Experimental Setup}
\label{sec:setup}

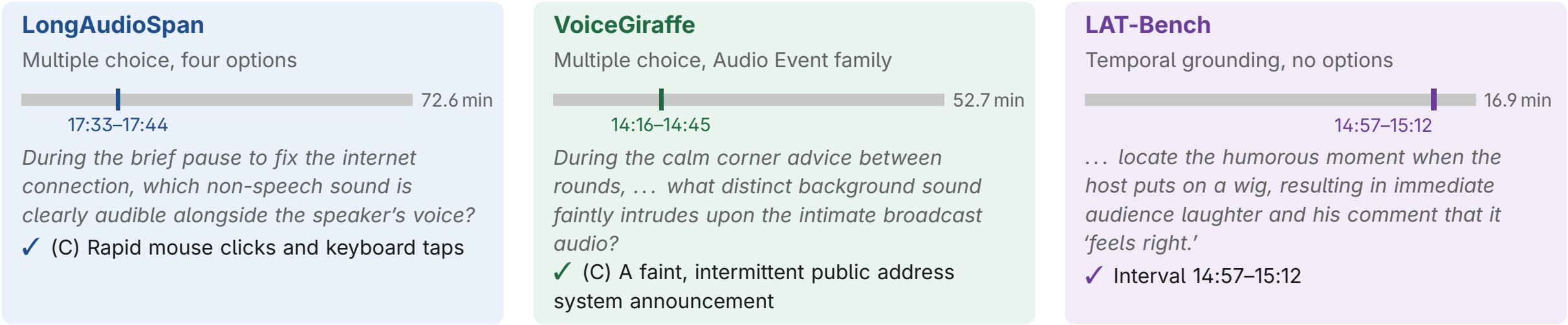
\begin{figure*}[t]
\centering
\begingroup\figfont\hyphenpenalty=10000\exhyphenpenalty=10000%
\definecolor{bmink}{gray}{0.10}%
\definecolor{bmmute}{gray}{0.38}%
\definecolor{bmrec}{gray}{0.78}%
\definecolor{bmA}{HTML}{1F4E8C}\definecolor{bmAt}{HTML}{EAF1FA}%
\definecolor{bmB}{HTML}{1E6B43}\definecolor{bmBt}{HTML}{E8F4ED}%
\definecolor{bmC}{HTML}{6A3D9A}\definecolor{bmCt}{HTML}{F1ECF7}%
\newcommand{\bmt}{\fontsize{8}{9.6}\selectfont}%
\newcommand{\bmf}{\fontsize{7}{8.4}\selectfont}%
\newcommand{\bms}{\fontsize{6.3}{7.4}\selectfont}%
\newcommand{\bmcard}[6]{%
  \fill[#3, rounded corners=3pt] (#1,0.32) rectangle (#1+#2,3.62);
  \node[anchor=base west, inner sep=0pt, font=\bmt\bfseries, text=#4] at (#1+0.2,3.30) {#5};
  \node[anchor=base west, inner sep=0pt, font=\bmf, text=bmmute] at (#1+0.2,2.95) {#6};}%
\newcommand{\bmrecord}[8][north]{%
  \fill[bmrec] (#2,2.56) rectangle (#2+#3,2.68);
  \node[anchor=west, inner sep=0pt, font=\bms, text=bmmute] at (#2+#3+0.08,2.62) {#4\,min};
  \pgfmathsetmacro{\bma}{#2+#3*#5/#4}\pgfmathsetmacro{\bmb}{max(#2+#3*#6/#4,\bma+0.045)}%
  \fill[#8] (\bma,2.51) rectangle (\bmb,2.73);
  \node[anchor=#1, inner sep=0pt, font=\bms, text=#8] at ({(\bma+\bmb)/2},2.44) {#7};}%
\begin{tikzpicture}[x=1cm, y=1cm,
  quote/.style={anchor=north west, inner sep=0pt, font=\bmf\itshape, text=bmmute, align=left, text width=4.73cm},
  answer/.style={anchor=north west, inner sep=0pt, font=\bmf, text=bmink, align=left, text width=4.73cm}]
\useasboundingbox (0,0.32) rectangle (16.0,3.62);

\bmcard{0}{5.13}{bmAt}{bmA}{LongAudioSpan}{Multiple choice, four options}
\bmrecord{0.2}{4.0}{72.6}{17.55}{17.733}{17:33--17:44}{bmA}
\node[quote] (qa) at (0.2,2.12) {During the brief pause to fix the internet connection, which non-speech sound is clearly audible alongside the speaker's voice?};
\node[answer] at ([yshift=-0.10cm]qa.south west) {\textcolor{bmA}{\ding{51}}\enspace(C) Rapid mouse clicks and keyboard taps};

\bmcard{5.435}{5.13}{bmBt}{bmB}{VoiceGiraffe}{Multiple choice, Audio Event family}
\bmrecord{5.635}{4.0}{52.7}{14.267}{14.75}{14:16--14:45}{bmB}
\node[quote] (qv) at (5.635,2.12) {During the calm corner advice between rounds, \ldots\ what distinct background sound faintly intrudes upon the intimate broadcast audio?};
\node[answer] at ([yshift=-0.10cm]qv.south west) {\textcolor{bmB}{\ding{51}}\enspace(C) A faint, intermittent public address system announcement};

\bmcard{10.87}{5.13}{bmCt}{bmC}{LAT-Bench}{Temporal grounding, no options}
\bmrecord[north east]{11.07}{4.0}{16.9}{14.95}{15.2}{14:57--15:12}{bmC}
\node[quote] (ql) at (11.07,2.12) {\ldots\ locate the humorous moment when the host puts on a wig, resulting in immediate audience laughter and his comment that it `feels right.'};
\node[answer] at ([yshift=-0.10cm]ql.south west) {\textcolor{bmC}{\ding{51}}\enspace Interval 14:57--15:12};
\end{tikzpicture}%
\endgroup
\caption{\textbf{The three benchmarks, with real questions.}
Each bar is a whole recording drawn to scale, and the mark on it is the annotated evidence.
Questions are shortened only with ellipses, and Appendix~\ref{app:examples} gives full examples.}
\label{fig:benchmarks}
\end{figure*}

Our experiments test whether the reader's scan finds the evidence for a question, and whether answering from the selected audio improves accuracy at a fixed audio budget.
The same reader answers from every audio and text input, so differences reflect only what it receives.

\subsection{Benchmarks}
\label{sec:benchmarks}

We evaluate on three benchmarks of long English and Mandarin recordings (Figure~\ref{fig:benchmarks}).

\paragraph{LongAudioSpan.}
LongAudioSpan \citep{huang2026longaudiospan} contains 360 recordings of 10 to 164 minutes, with three native questions written from each unedited recording.
We evaluate its two multiple-choice modes.
The \emph{accuracy} mode asks the 1{,}080 native questions with four options each.
The \emph{chain} mode asks three linked questions about an edited copy of the recording with one to three planted edits of one type.
An edit is an inserted sound event, a stretch replaced by silence or masked by noise, or a change of playback speed or volume.
Q1 names the edited event and asks how many occurrences there are and where the $k$-th one is, and its four options cross two counts with two times.
Q2 then asks what was said at that point, and Q3 asks how that relates to the rest of the recording.
A chain scores 0, 33.3, 66.7 or 100 for zero to three consecutive correct answers from Q1, and random guessing gives 10.9 in expectation.
We split the recordings into 180 development and 180 test recordings, with 540 native questions and 180 occurrence chains per half.
This split is stratified by edit type, language and duration, and the test recordings average 48.3 minutes.
We choose the scan's use of answer options and its number of windows on development recordings.
The same recordings choose the detector recipes, candidate thresholds and calibrators of the occurrence extension.
We then evaluate the fixed configuration once on the test half in each mode.

\paragraph{VoiceGiraffe.}
VoiceGiraffe \citep{ye2026voicegiraffe} contains 123 recordings of 35 to 79 minutes, and we evaluate 1{,}499 of its 1{,}500 multiple-choice questions.
We call its Audio Event and Paralinguistic families, which ask about non-speech sounds and voice properties, the \emph{sound families} (549 questions).
The other four concern the content (950 questions).

\paragraph{LAT-Bench.}
LAT-Bench \citep{shao2026listeningwithtime} provides 1{,}002 temporal grounding queries about 247 recordings of up to 30 minutes.
Each query describes a moment by its speech, music and sound effects, and a system predicts its start and end times.
We report the mean temporal IoU (mIoU) and the recall at IoU thresholds of 0.3, 0.5 and 0.7, averaged equally over English and Chinese.

\subsection{Reader and Metrics}
\label{sec:setup-data}

\paragraph{Reader.}
We use Qwen3-Omni-30B-A3B-Instruct \citep{xu2025qwen3omni} as the reader, with frozen weights and greedy decoding.
The reader receives 16\,kHz waveforms, and each selected excerpt is cropped from the waveform before the reader's audio encoder.
VoiceGiraffe and LAT-Bench use the LongAudioSpan prompts, windows and budgets without change (Appendix~\ref{app:repro}).

\paragraph{Metrics.}
The \emph{evidence hit rate} is the percentage of annotated questions for which any selected excerpt overlaps any annotated evidence range, even partially.
We use the annotated support ranges on LongAudioSpan and the evidence times on VoiceGiraffe.
For question answering, we report multiple-choice accuracy.
A \emph{time-option question} has a clock time in at least three of its four options.
Intervals are 95\% bootstrap confidence intervals that resample recordings, since questions about the same recording are not independent.

\paragraph{Efficiency.}
Since the scan depends on the question, we time one question per recording.
The timed questions come from 22 test LongAudioSpan recordings, 10 VoiceGiraffe recordings and 10 LAT-Bench recordings.
Each time runs from decoding the audio file to the answer (Appendix~\ref{app:repro}).

\subsection{Baselines}
\label{sec:baselines}

\paragraph{Selection baselines.}
These baselines give the reader the same 60 seconds of audio as ReListen but choose it differently.
\emph{Uniform sampling} takes 20-second windows centred at one sixth, one half and five sixths of the recording.
Our \emph{learned audio retriever} is a scorer trained on frozen reader features, which selects eight short regions.
\emph{BM25} \citep{robertson2009bm25} and \emph{BGE-M3} \citep{chen2024bgem3} rank 30-second transcript chunks by word matching and dense-embedding similarity, respectively.
Both use the question and its options as the query and select the audio of the top two chunks.

\paragraph{Input baselines.}
These baselines change what the same reader receives.
\emph{No audio} supplies the question and its answer options alone.
\emph{Full recording} supplies the audio up to an 80-minute cap, which covers every VoiceGiraffe recording.
Its \emph{time marks} variant precedes each consecutive 30-second segment with its start time as text.
\emph{Whisper transcript} and \emph{reader transcript} supply the text of the whole recording, which Whisper large-v3 \citep{radford2023whisper} or the reader transcribes in consecutive 30-second excerpts.
They carry no time marks and no audio, so they measure answering from the words alone.
Finally, \emph{annotated evidence} serves as a reference rather than a baseline.
It supplies the evidence intervals, padded to at least 20 seconds each and capped at 120 seconds in total.

\paragraph{Prior systems.}
We run the released LAT-Audio model \citep{shao2026listeningwithtime} with its official tool policy on the same LAT-Bench queries as ReListen.
Echo \citep{wu2026echo} reasons over the audio itself, but its trained checkpoint is unavailable.
Our \emph{Echo-style} system therefore gives the frozen reader its whole timestamped transcript and lets it request audio while reasoning.
\emph{Speaker-Reasoner} \citep{lin2026speakerreasoner} transcribes each recording with its released Stage 2 model, and the reader answers from this timestamped transcript (Appendix~\ref{app:prior-art}).
We call the transcript inputs and these two systems \emph{text-mediated}, since they convert the whole recording to text before any question is answered.

\section{Results}
\label{sec:results}

\begin{table*}[t]
\centering
\small
\setlength{\tabcolsep}{5.5pt}
\begin{tabular}{lccrcrccc}
\toprule
 & & & \multicolumn{2}{c}{LongAudioSpan} & \multicolumn{4}{c}{VoiceGiraffe} \\
\cmidrule(lr){4-5}\cmidrule(lr){6-9}
Method & Training & GiB & s & Acc. & s & All & Sound & Content \\
\midrule
\grouprow{9}{Text-mediated}
Reader transcript & \ding{55} & 71.9 & 119 & \textbf{59.6} & 135 & 52.8 & 50.8 & 54.0 \\
Whisper transcript & \ding{55} & 68.4 & 27.6 & 58.1 & 28.4 & -- & -- & -- \\
Echo-style \citep{wu2026echo} & \ding{55} & 72.8 & 157 & 58.1 & 169 & 58.1 & 51.0 & 62.2 \\
Speaker-Reasoner \citep{lin2026speakerreasoner} & \ding{51} & 78.1 & 1124 & 53.0 & 1048 & 55.7 & 52.6 & 57.5 \\
\midrule
\grouprow{9}{Audio input}
No audio & \ding{55} & -- & -- & 43.5 & -- & 34.5 & 37.3 & 32.8 \\
Uniform excerpts & \ding{55} & -- & -- & 45.0 & -- & 40.7 & 48.5 & 36.2 \\
Full recording & \ding{55} & 78.2 & 11.4 & 54.6 & 11.6 & 51.4 & 51.4 & 51.5 \\
ReListen (ours) & \ding{55} & \textbf{67.8} & 17.3 & 53.5 & 17.0 & \textbf{62.0} & \textbf{58.7} & \textbf{64.0} \\
\quad four GPUs & & \textbf{67.8} & 5.6 & & 5.7 & & & \\
ReListen-Fast (ours) & \ding{55} & \textbf{67.8} & 10.0 & 53.7 & 9.0 & 60.9 & 57.7 & 62.7 \\
\quad four GPUs & & \textbf{67.8} & \textbf{3.7} & & \textbf{3.6} & & & \\
\bottomrule
\end{tabular}
\caption{\textbf{Accuracy and cost per question.}
All rows use the same frozen reader to answer, and we report accuracy (\%) on test LongAudioSpan and on VoiceGiraffe.
Seconds (s) are means over the timed questions.
GiB is the peak memory on any one GPU, including the reader's 59\,GiB of weights, and Training marks a model fine-tuned for the task.
Uniform excerpts and ReListen give the reader 60 seconds of audio, and the full recording carries time marks.
Rows use one GPU unless marked four GPUs, which reproduce the one-GPU selections and accuracy.
The full recording uses two GPUs beyond about 62 minutes.}
\label{tab:summary}
\end{table*}

\subsection{Experimental Results}
\label{sec:content-results}

We report multiple-choice accuracy and cost per question on LongAudioSpan and VoiceGiraffe in Table~\ref{tab:summary}.
Appendices~\ref{app:content} and~\ref{app:voicegiraffe} report results by dimension, duration and question family.

\begin{figure}[t]
\centering
\begingroup\figfont%
\begin{tikzpicture}
\begin{groupplot}[group style={group size=2 by 1, horizontal sep=0.42cm, y descriptions at=edge left},
  vizaxis, scale only axis, width=2.84cm, height=3.4cm, xmode=log, log basis x=10, xmin=40, xmax=5500, ymin=35, ymax=78,
  xtick={60,600,3600}, xticklabels={1 min,10 min,1 h}, ytick={40,50,60,70},
  legend columns=4, legend style={legend image post style={scale=0.7}, /tikz/every even column/.append style={column sep=3pt}}]
\nextgroupplot[title={LongAudioSpan}, ylabel={Accuracy (\%)}, legend to name=efflegend]
\addplot[vizorange, vizline] table[x=seconds, y=uniform] {figures/efficiency_las.dat}; \addlegendentry{Uniform}
\addplot[vizblue, vizline] table[x=seconds, y=relisten] {figures/efficiency_las.dat}; \addlegendentry{ReListen}
\addplot[black!80, only marks, mark=square*, mark size=1.3pt] coordinates {(2650,54.6)}; \addlegendentry{Full recording}
\addplot[black!80, only marks, mark=diamond*, mark size=1.8pt] coordinates {(69,72.7)}; \addlegendentry{Annotated evidence}
\nextgroupplot[title={VoiceGiraffe}]
\addplot[vizorange, vizline, only marks] table[x=seconds, y=uniform] {figures/efficiency_vg.dat};
\addplot[vizblue, vizline, only marks] table[x=seconds, y=relisten] {figures/efficiency_vg.dat};
\addplot[black!80, only marks, mark=square*, mark size=1.3pt] coordinates {(3240,51.4)};
\addplot[black!80, only marks, mark=diamond*, mark size=1.8pt] coordinates {(57,74.5)};
\end{groupplot}
\node[font=\fontsize{7}{8}\selectfont, text=vizink, anchor=north] (xl) at ($(group c1r1.south east)!0.5!(group c2r1.south west)+(0,-0.32)$) {Audio given to the reader};
\node[anchor=north] at ($(xl.south)+(0,0.06)$) {\pgfplotslegendfromname{efflegend}};
\end{tikzpicture}%
\endgroup
\caption{\textbf{Accuracy against the audio given to the reader.}
Multiple-choice accuracy on test LongAudioSpan and on VoiceGiraffe.
The full recording is placed at its mean duration up to 80 minutes, and the annotated evidence at its mean length.}
\label{fig:efficiency}
\end{figure}
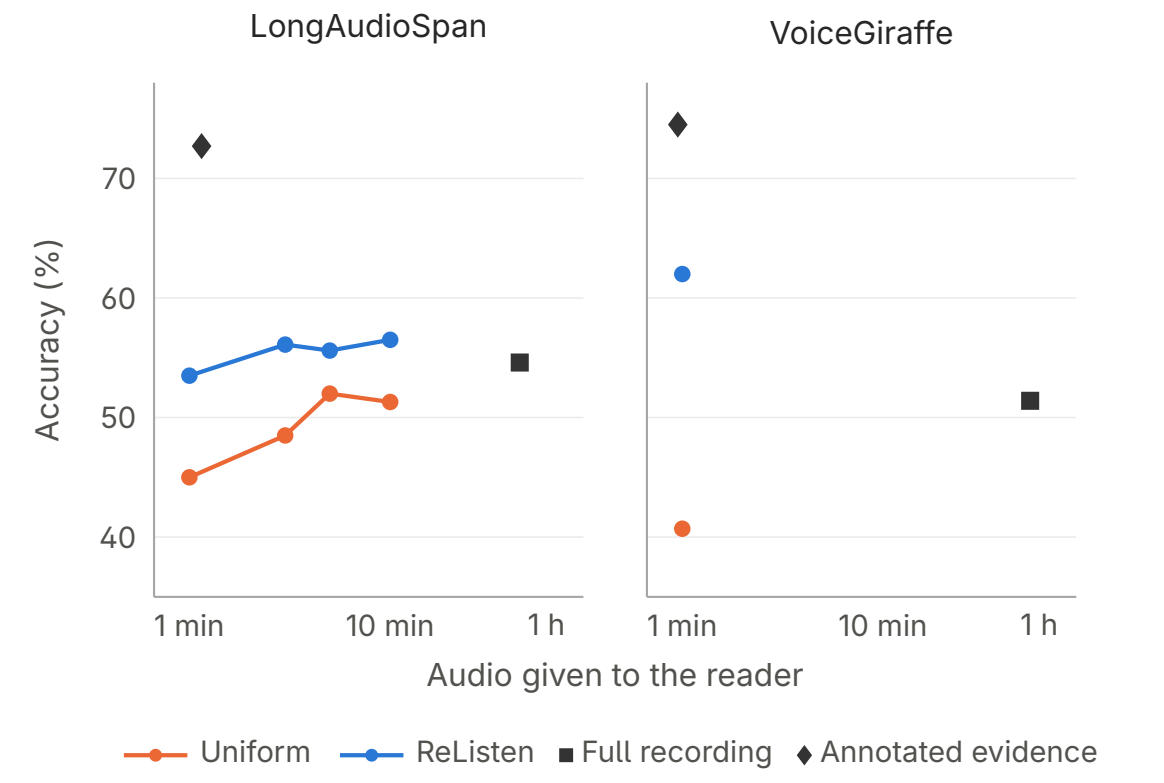

\paragraph{Overall Task Performance.}
With the same 60-second budget, our ReListen outperforms uniform excerpts by 8.5 points on LongAudioSpan and 21.3 points on VoiceGiraffe.
Compared with the full recording, which gives the reader 44 to 54 times more audio, ReListen performs competitively on LongAudioSpan and outperforms it by 10.6 points on VoiceGiraffe (Figure~\ref{fig:efficiency}).
In summary, ReListen achieves the best or competitive accuracy among audio inputs, which verifies that a frozen reader can select the evidence it needs to answer.

\paragraph{Comparison with Transcripts.}
On VoiceGiraffe, ReListen outperforms the reader's transcript of the whole recording by 9.2 points.
The Echo-style and Speaker-Reasoner systems, which read timestamped transcripts, also perform below ReListen (Appendix~\ref{app:prior-art}).
On LongAudioSpan, whose questions mostly concern what is said, the transcript performs better on questions without time options.

\subsection{Inference Efficiency}
\label{sec:cost}

\begin{figure}[t]
\centering
\input{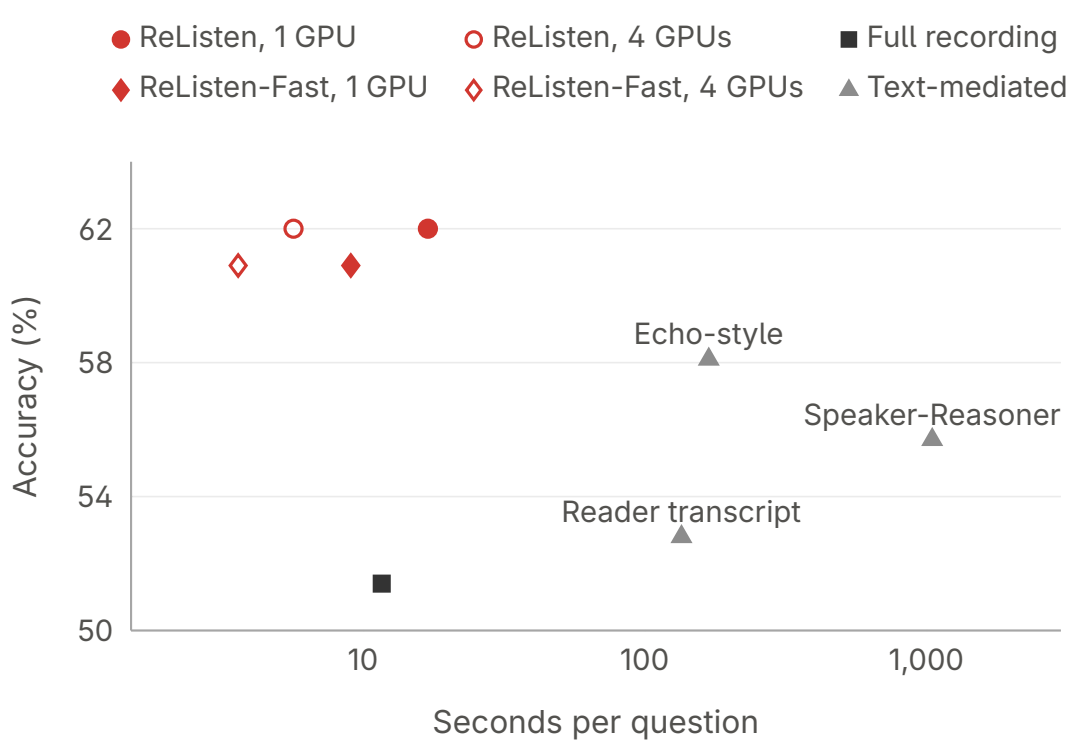}%
\begingroup\figfont%
\pgfplotsset{atmark/.style={only marks, mark size=1.6pt, line width=0.6pt},
  atref/.style={atmark, black!45, mark=triangle*, mark size=1.8pt}}
\tikzset{atlab/.style={font=\fontsize{6.5}{7}\selectfont, text=vizink, inner sep=1.5pt}}
\begin{tikzpicture}
\begin{axis}[vizaxis, scale only axis, width=6.15cm, height=3.1cm, xmode=log, log basis x=10, xmin=1.5, xmax=3000,
  xtick={10,100,1000}, xticklabels={10,100,{1,000}}, ymin=50, ymax=64, ytick={50,54,58,62},
  xlabel={Seconds per question}, ylabel={Accuracy (\%)},
  legend columns=3, legend to name=atlegend, legend cell align=left,
  legend style={legend image post style={scale=0.9}, /tikz/every even column/.append style={column sep=5pt}}]
\addplot[atmark, vizred, mark=*] coordinates {(\atVGReListenX,\atVGReListenY)}; \addlegendentry{ReListen, 1 GPU}
\addplot[atmark, vizred, mark=o, mark options={fill=white}] coordinates {(\atVGReListenFourX,\atVGReListenFourY)}; \addlegendentry{ReListen, 4 GPUs}
\addplot[atmark, black!80, mark=square*, mark size=1.4pt] coordinates {(\atVGFullX,\atVGFullY)}; \addlegendentry{Full recording}
\addplot[atmark, vizred, mark=diamond*, mark size=2pt] coordinates {(\atVGFastX,\atVGFastY)}; \addlegendentry{ReListen-Fast, 1 GPU}
\addplot[atmark, vizred, mark=diamond, mark size=2pt, mark options={fill=white}] coordinates {(\atVGFastFourX,\atVGFastFourY)}; \addlegendentry{ReListen-Fast, 4 GPUs}
\addplot[atref] coordinates {(\atVGReaderTranscriptX,\atVGReaderTranscriptY) (\atVGEchoX,\atVGEchoY) (\atVGSpeakerX,\atVGSpeakerY)}; \addlegendentry{Text-mediated}
\node[atlab, anchor=south] at (axis cs:\atVGReaderTranscriptX,\atVGReaderTranscriptY) {Reader transcript};
\node[atlab, anchor=south] at (axis cs:\atVGEchoX,\atVGEchoY) {Echo-style};
\node[atlab, anchor=south] at (axis cs:\atVGSpeakerX,\atVGSpeakerY) {Speaker-Reasoner};
\end{axis}
\node[anchor=south] at ($(current axis.north)+(0,0.12)$) {\pgfplotslegendfromname{atlegend}};
\end{tikzpicture}%
\endgroup
\caption{\textbf{Accuracy against seconds per question on VoiceGiraffe.}
Each point gives a method's mean time over the timed questions and its test accuracy (Table~\ref{tab:summary}).
Filled marks use one GPU, and open marks divide the scan over four GPUs with the same selection.
The horizontal axis is logarithmic, and every ReListen point lies left of all text-mediated systems.}
\label{fig:latency}
\end{figure}

Table~\ref{tab:summary} also reports the mean seconds per question, and Figure~\ref{fig:latency} plots accuracy against them on VoiceGiraffe.

\paragraph{Comparison with Text-Mediated Systems.}
On LongAudioSpan, ReListen answers a question in 17.3 seconds on one GPU.
In contrast, the text-mediated systems take 27.6 to 1124 seconds, since they first transcribe the whole recording.
On both benchmarks, ReListen is seven to eight times faster than the reader's transcript pipeline.
Therefore, ReListen is faster than every text-mediated system and, on VoiceGiraffe, also more accurate.

\paragraph{Comparison with Trained Grounding.}
The trained LAT-Audio model grounds a LAT-Bench query in 173 seconds on average.
In contrast, the scan grounds the same queries in 3.1 seconds without any training, at a comparable mIoU (Table~\ref{tab:grounding}).

\begin{table}[t]
\centering
\small
\setlength{\tabcolsep}{4pt}
\begin{tabular}{lccc}
\toprule
 & LAS & \multicolumn{2}{c}{VoiceGiraffe} \\
\cmidrule(lr){2-2}\cmidrule(l){3-4}
Method & All & All & Sound \\
\midrule
Uniform sampling & 18.7 & 11.1 & 4.4 \\
Learned audio retriever & 32.6 & 16.5 & 5.8 \\
BM25 & 45.2 & 58.8 & 45.6 \\
BGE-M3 & 47.8 & 52.1 & 37.6 \\
\midrule
ReListen (ours) & \textbf{69.2} & \textbf{88.5} & \textbf{76.5} \\
\bottomrule
\end{tabular}
\caption{\textbf{Evidence localization with a 60-second budget.}
The hit rate (\%) is the fraction of questions for which any selected excerpt overlaps the annotated evidence.
LAS is test LongAudioSpan.}
\label{tab:localization}
\end{table}

\paragraph{Comparison with the Full Recording.}
On one GPU, the full recording is faster than ReListen-Fast only for recordings under about 30 minutes.
Its time grows faster than linearly with duration, and it exceeds one 80\,GB GPU beyond about 62 minutes.
In contrast, the time of the scan grows linearly, and its memory stays below 68\,GiB at every duration.

\paragraph{Parallel Scan and ReListen-Fast.}
Since the reader scores every window independently, the scan divides across GPUs.
On the two benchmarks, four GPUs make ReListen 3.1 and 3.0 times faster than one GPU.
It is then faster than the full recording on 29 of the 32 timed questions.
As an efficiency setting, \emph{ReListen-Fast} uses a hop of 20 seconds, so that the scan hears each second once.
Compared with ReListen, its accuracy changes by $+0.2$ points on LongAudioSpan and $-1.1$ on VoiceGiraffe, and both 95\% intervals include zero.
We keep the 10-second hop, chosen on development recordings, as the main configuration.
In summary, ReListen is faster than every text-mediated system at constant memory, which verifies that searching the audio itself is efficient.

\subsection{Evidence Localization}
\label{sec:locate-results}

This section investigates whether the scan selects the evidence that a question needs.
To this end, we measure how often each 60-second selection overlaps the annotated evidence (Table~\ref{tab:localization}).

\paragraph{Comparison with Retrieval.}
The scan hits the evidence for 69.2\% of LongAudioSpan and 88.5\% of VoiceGiraffe questions, whereas transcript retrieval hits it for at most 47.8\% and 58.8\%.
The gap is largest on sound questions, where transcript retrieval falls to 45.6\% and the scan keeps 76.5\%, since sounds and voices leave no words to match.

\begin{table}[t]
\centering
\small
\setlength{\tabcolsep}{3.2pt}
\begin{tabular}{lcccc}
\toprule
Method & mIoU & R@0.3 & R@0.5 & R@0.7 \\
\midrule
Uniform, centre window & 2.9 & 3.8 & 1.9 & 0.8 \\
BM25, best 30-s chunk & 34.2 & 56.6 & 29.4 & 11.1 \\
Full recording & 12.3 & 17.9 & 10.0 & 5.1 \\
\quad with time marks & 39.3 & 61.9 & 37.7 & 15.7 \\
Gemini-2.5-Pro$^\dagger$ & 42.8 & 63.3 & \textbf{51.3} & 26.9 \\
LAT-Audio & 46.3 & 62.0 & 48.4 & \textbf{32.5} \\
\midrule
ReListen (ours) & \textbf{47.7} & \textbf{77.2} & 50.0 & 19.2 \\
\bottomrule
\end{tabular}
\caption{\textbf{Event grounding on LAT-Bench.}
We report mean IoU and recall at IoU 0.3, 0.5 and 0.7 (\%).
ReListen returns the expanded region of its scan.
Appendix~\ref{app:voicegiraffe} gives ablations.
$^\dagger$Reported by \citet{shao2026listeningwithtime}.}
\label{tab:grounding}
\end{table}

\paragraph{Event Grounding.}
On LAT-Bench, the expanded region of the scan reaches 47.7 mIoU without any change, which is comparable to the trained LAT-Audio model on the same queries (Table~\ref{tab:grounding}).
Its recall at an IoU of 0.7 is lower, so the scan finds the right region with less precise boundaries.

\subsection{Listening versus Reading}
\label{sec:reading-results}

\begin{table}[t]
\centering
\small
\setlength{\tabcolsep}{2.2pt}
\begin{tabular}{lcccccc}
\toprule
 & \multicolumn{2}{c}{LAS} & \multicolumn{2}{c}{VG content} & \multicolumn{2}{c}{VG sound} \\
\cmidrule(lr){2-3}\cmidrule(lr){4-5}\cmidrule(lr){6-7}
Excerpt source & Audio & Text & Audio & Text & Audio & Text \\
\midrule
ReListen, 60 s & 60.5 & 59.4 & 64.0 & 65.2 & 58.7 & 50.1 \\
Annotated evidence & 70.9 & 70.1 & 78.3 & 76.3 & 67.9 & 48.4 \\
\bottomrule
\end{tabular}
\caption{\textbf{Listening versus reading the same excerpts.}
The reader hears the excerpts (Audio) or reads its own transcript of them (Text), and we report accuracy (\%).
LAS uses test LongAudioSpan questions without time options, with audio sliced from whole-recording encoder outputs.}
\label{tab:matched}
\end{table}

This section further investigates whether the selected evidence is worth hearing rather than reading.
To this end, we let the reader read its own transcript of exactly the excerpts it would hear (Table~\ref{tab:matched}).
On content questions, listening and reading differ by about two points or less, for both the scan's excerpts and the annotated evidence.
Therefore, the lead of a whole transcript on LongAudioSpan comes from covering the recording rather than from the text.
On VoiceGiraffe sound questions, however, listening outperforms reading by 8.6 points on the scan's excerpts and by 19.5 points on the annotated evidence.
This verifies our motivation to answer from the audio itself, since a transcript discards the sounds and voices that these questions ask about.

\subsection{Upper Bound with Annotated Evidence}
\label{sec:gap-results}

To analyze the remaining errors, we replace the selected minute by the annotated evidence.
The reader then answers 72.7\% on LongAudioSpan and 74.5\% on VoiceGiraffe, 19.2 and 12.5 points above ReListen (Tables~\ref{tab:native} and~\ref{tab:voicegiraffe}).
On questions without time options, the annotated evidence also outperforms the whole transcript, 73.0\% against 65.7\% on LongAudioSpan and 69.7\% against 55.7\% on VoiceGiraffe.
On LongAudioSpan, when the scan misses the evidence, ReListen answers 38.7\%, the same as without audio, and these misses account for half of the gap.
When the scan hits the evidence, more audio does not help, since one, three and ten minutes yield about 64\% on questions without time options (Appendix~\ref{app:matched}).
These results suggest that more precise selection, rather than more audio, would further improve ReListen.

\subsection{Questions about the Whole Recording}
\label{sec:global-results}

However, more audio does help when the evidence spreads across a recording.
On 61 test questions of LongAudioSpan, the annotated evidence extends over at least half of the recording.
There, the selected minute answers 55.7\%, against 73.8\% for the full recording.
Ten minutes from the same scan reach 72.1\%, although a gap remains on recordings of 30 to 60 minutes (Appendix~\ref{app:global}).
Beyond ten minutes, the budget can grow up to the whole recording that the full-recording baseline hears.
On VoiceGiraffe, ReListen outperforms the full recording on the 31 such questions, 54.8\% against 35.5\%.
For questions about the $k$-th occurrence of an event, the occurrence extension instead weighs candidates from the whole recording (Section~\ref{sec:occurrence}).
In summary, the same scan with a larger budget closes most of the gap to the full recording, which verifies that the budget rather than the scan limits ReListen on these questions.

\subsection{Occurrence Questions}
\label{sec:chain-results}

Occurrence chains ask for the $k$-th occurrence of an event, and every occurrence matches the event's description equally well.
Hence, selection by relevance alone cannot tell the occurrences apart, and the Whisper transcript reaches 10.2, below random guessing at 10.9 (Table~\ref{tab:mainfull}).
Even the full recording with time marks reaches only 17.2.
Our supervised extension reaches a chain score of 60.7 and outperforms verifying retrieved candidates one by one, which reaches 37.4.
Most of this gain comes from the detectors, since the scan alone reaches 35.9 and the detectors alone 52.0.
A thresholded count instead of the occurrence estimate lowers the score to 55.9 (Appendix~\ref{app:ablation}).
For coughs in meetings, which have no detector, the scan alone places the $k$-th cough within 30 seconds for 37.0\% of the questions, against 3.8\% for a random time (Appendix~\ref{app:natural}).

\section{Conclusion}

A frozen audio language model can locate the evidence for a question in a long recording without training a selector.
Its selected minute overlaps the annotated evidence for 69.2\% of test LongAudioSpan and 88.5\% of VoiceGiraffe questions.
Answering from it matches or exceeds hearing the whole recording with 44 to 54 times less audio.
On VoiceGiraffe the selected minute exceeds a transcript of the whole recording.
On sound questions, selected audio also exceeds a transcript of the same excerpts.
It also answers faster than every transcript-based system, with constant memory and a scan that divides across GPUs.
The remaining gap to the annotated evidence calls for selection that isolates the needed sound more precisely.

\section*{Limitations}

Our evidence comes from one reader and three benchmarks of mostly English and Mandarin speech.
Transfer to other model families, music, sound scenes or other languages is therefore untested.
Our cost comparison also assumes one question per recording, whereas with several questions a transcript can be reused while the scan reruns for each question.
ReListen also uses one budget for every question, and choosing the budget for each question is untested.
Questions that need more audio than the reader can hear at once, such as summaries of very long recordings, lie beyond this work.

\section*{Ethical Considerations}

Evidence localization can help users search meetings, calls and other long recordings, but can also expose statements from private conversations without the speakers' consent, so applications should respect consent and access restrictions for both recordings and their transcripts.
All three benchmarks are research releases of human speech, VoiceGiraffe waveforms were obtained under its gated access agreement, and we redistribute no recordings.
Some questions concern age, gender or emotion inferred from a voice, which we report as benchmark categories rather than proposed applications, and since the systems still make substantial errors, their outputs should not be used for decisions about individuals.

\bibliography{refs}
\appendix
\section{Benchmark Examples}
\label{app:examples}

\newcommand{\examplebox}[2]{\par\noindent\fbox{\parbox{\dimexpr\columnwidth-2\fboxsep-2\fboxrule\relax}{\small\textbf{#1}\par\smallskip #2}}\par\medskip}

We show examples from the three benchmarks.
The correct option is in bold, and the evidence is the annotated time range.

\subsection{LongAudioSpan}

LongAudioSpan assigns each native question one of three levels, perception, understanding and reasoning, and one recording receives one question per level.
The first three boxes show one question of each level in the accuracy mode.
Note that a perception question can already require searching a long recording: in the first example, the 11-second evidence lies 17 minutes into a 73-minute recording.

\examplebox{Accuracy mode, perception (recognition). 72.6-minute recording.}{%
During the brief pause to fix the internet connection, which non-speech sound is clearly audible alongside the speaker's voice?\par
(A) Heavy breathing from exertion\quad
(B) A loud digital static burst\quad
\textbf{(C) Rapid mouse clicks and keyboard taps}\quad
(D) Distant traffic noise outside\par
\emph{Evidence:} 17:33--17:44.}

\examplebox{Accuracy mode, understanding (tracking). 65.1-minute recording.}{%
At what point does the primary geographic focus of the festival broadcast shift from Colorado to Australia?\par
(A) 05:13--05:45\quad
\textbf{(B) 34:41--36:29}\quad
(C) 32:52--34:41\quad
(D) 44:45--48:35\par
\emph{Evidence:} 34:41--36:29.}

\examplebox{Accuracy mode, reasoning (relational). 10.5-minute recording.}{%
What unstated connection do the commentators draw between the Bulgarian pair's earlier pedigree and their later play?\par
\textbf{(A) It gives them resilience, but not enough attacking variety}\par
(B) It proves they are coasting on reputation alone\par
(C) It makes them the sharper pair at the net\par
(D) It shows their coaching is driving the late tactical changes\par
\emph{Evidence:} 05:37--05:53 and 10:00--10:25.}

Occurrence chains are asked about an edited copy of a recording.
In the example below, the same 6.5-second clip of bird song is inserted twice, at 09:36 and at 30:09 of a 40.8-minute recording.
Q1 asks for the count and the position of one occurrence.
In contrast, Q2 and Q3 refer to that occurrence without stating its time, so they can be answered only after it is found.

\examplebox{Occurrence chain, inserted sound event. 40.8-minute edited recording.}{%
\emph{Q1.} This audio contains bird singing. How many instances are there in total, and approximately where does the 1st one occur?\par
\textbf{(A) 2 in total, the 1st near 09:39}\quad
(B) 2 in total, the 1st near 30:12\par
(C) 1 in total, the 1st near 09:39\quad
(D) 1 in total, the 1st near 30:12\par\smallskip
\emph{Q2.} This audio contains bird singing. At the location of the first occurrence, what is being discussed?\par
\textbf{(A) A male host rapidly lists specific Breeze Copilot capabilities like CRM object updates and engagement summaries.}\par
(B) A female narrator defines the Commerce Analytics Suite as a tool for streamlining sales billing and payments.\par
(C) A discussion focuses on editing video clip transcripts to improve the workflow for creating short content.\par
(D) Speakers reflect on breaking down silos to create a unified view across the entire customer journey.\par\smallskip
\emph{Q3.} This audio contains bird singing. How does the content at the location of the first occurrence relate to other parts of the audio?\par
(A) The content here serves as a thematic callback to the subsequent demonstration of using Claude for project management tasks.\par
(B) The content here provides a comparative example to the later discussion on external AI models integrating via MCP.\par
(C) The content here acts as a verification check that contradicts the earlier announcement of the Commerce Analytics Suite availability.\par
\textbf{(D) The content here forms a feature expansion relation to the later section on collaborative workflow building with Copilot.}}

The other edit types use the same three templates.
For example, a Q1 about a speed edit reads ``This audio contains sudden playback speedup. How many instances are there in total, and approximately where does the 2nd one occur?''

\subsection{VoiceGiraffe}

VoiceGiraffe groups its questions into six families.
The two sound families ask about non-speech sounds and about voice properties, and the Paralinguistic family covers emotion, pitch, timbre, age and gender.
In contrast, the four content families ask about what is said.
They differ in their options: time ranges for Temporal Localization, orderings of numbered events for Causal Alignment, and counts or outcomes across a long stretch for Event Tracking.
Many questions first describe a scene in words and then ask about it, so the description serves to find the evidence and the question tests what is heard there.

\examplebox{Audio Event. 52.7-minute recording.}{%
During the calm corner advice between rounds, as the coaches instruct their fighter to breathe, ``catch some air,'' and prevent his opponent from getting comfortable, what distinct background sound faintly intrudes upon the intimate broadcast audio?\par
(A) A low-volume, rhythmic electronic arena music beat\par
(B) The distant, metallic clanking of a ringside warning bell\par
\textbf{(C) A faint, intermittent public address system announcement}\par
(D) An organized, rhythmic crowd chant of the challenger's name\par
\emph{Evidence:} 14:16--14:45.}

\examplebox{Paralinguistic (emotion). 60.3-minute recording.}{%
During the plot twist reveal about the successful audition, how does the friend's vocal delivery of the seemingly supportive phrase, ``Well, that's good. Because you didn't. And I'm incredibly happy for you,'' reveal his true emotional subtext?\par
(A) He employs a warm, sincere tone, expressing genuine forgiveness and immediate relief after the cruel prank.\par
\textbf{(B) He employs a loud, sarcastic tone, executing a reverse fake-out to playfully retaliate against his friend.}\par
(C) He employs a breathless, panicked tone, revealing lingering anxiety over the initially misunderstood audition message.\par
(D) He employs a soft, mournful tone, indicating deep internal resentment despite the outwardly supportive spoken words.\par
\emph{Evidence:} 57:58--58:00.}

\examplebox{Semantic Content. 59.9-minute recording.}{%
During the mock game show challenge immediately following the playful corporate recruiting dispute, what specific physical impairments does the distressed competitor explicitly report experiencing while attempting the task?\par
(A) Triggering electrical buzzer penalties and dropping the extracted target components into the restricted cavity.\par
(B) Losing grip stability on the required hand tools and slipping repeatedly on accidentally spilled chemical fluids.\par
\textbf{(C) Swallowing stray paint and failing to operate the interface buttons due to the bulkiness of protective gloves.}\par
(D) Experiencing severe physical dehydration and suffering visual hallucinations due to prolonged bodily exhaustion.\par
\emph{Evidence:} 24:24--24:47.}

\examplebox{Temporal Localization. 68.0-minute recording. Options are in seconds.}{%
During the broader discussion analyzing how algorithmic social media captures the mind and isolates users from nuanced conversation, at what point do the speakers specifically draw a parallel between phone app engineering and slot-machine psychology?\par
(A) [2514, 2544]\quad
(B) [1289, 1315]\quad
(C) [2452, 2482]\quad
\textbf{(D) [1338, 1366]}\par
\emph{Evidence:} 22:18--22:46.}

\examplebox{Causal Alignment. 67.4-minute recording.}{%
During the chaotic overtime phase on the second map, how does FURIA successfully establish numerical superiority and completely sweep the defensive site?\par
{[1]} A chaotic overtime scramble ensues, characterized by deep flank hunts.\par
{[2]} Vitality safely preserves their expensive sniper weaponry by retreating off-site.\par
{[3]} FURIA utilizes a deceptive smoke grenade fake to bait out the defenders.\par
{[4]} FURIA executes a lightning-fast, highly aggressive pathing maneuver.\par
{[5]} Vitality's superstar players fail to firmly lock down the defense.\par
{[6]} FURIA establishes numerical superiority and ruthlessly sweeps the site.\par
(A) [4]$\rightarrow$[5]$\rightarrow$[1]$\rightarrow$[6]\quad
(B) [5]$\rightarrow$[3]$\rightarrow$[4]$\rightarrow$[6]\par
\textbf{(C) [5]$\rightarrow$[1]$\rightarrow$[4]$\rightarrow$[6]}\quad
(D) [2]$\rightarrow$[1]$\rightarrow$[4]$\rightarrow$[6]\par
\emph{Evidence:} 41:05--42:38.}

\examplebox{Event Tracking. 47.0-minute recording.}{%
In the early portion of the audio, during the characters' extended debate over video game consoles and entertainment center space, how many specific past consumer devices are explicitly named as purchasing blunders or deep regrets?\par
(A) 4 devices\quad
(B) 1 device\quad
(C) 2 devices\quad
\textbf{(D) 3 devices}\par
\emph{Evidence:} 01:54--04:14.}

\subsection{LAT-Bench}

Finally, a LAT-Bench query describes one moment by its speech, its speaking style and the surrounding sounds, and asks for its interval without options.
The prediction is compared with the annotated interval by intersection over union.

\examplebox{Temporal grounding query. 2.7-minute recording.}{%
Please listen to the audio carefully and locate the first instance where the male speaker, in a guiding tone, provides instructions on how to handle the net immediately after the distinct sound of it splashing into the water. His speech begins with praise and then directs the woman to let the net sink before pulling it. Please strictly output the result in the format of [Start Time - End Time]. Do not output any extra explanatory text.\par
\emph{Annotated interval:} 01:09--01:26.}

\section{Prompts}
\label{app:prompts}

\paragraph{Jev-Style Scan.}
English: ``Question: \{question\}'', the four options on separate lines, a blank line, and ``Does this audio excerpt contain information that helps answer the question? Answer yes or no.''
The question-only variant omits the options.
The Mandarin prompt is a direct translation with a one-character answer.

\paragraph{Presence.}
English: ``Does this audio excerpt contain \{event\}? Answer yes or no.''
The Mandarin prompt is its direct translation and asks for a one-character answer.
Here, the event phrase is the noun phrase that follows ``This audio contains'' in the Q1 stem.
The score sums the probabilities of ``yes'', ``Yes'' and the Mandarin affirmative, and of ``no'', ``No'' and the two Mandarin negatives.

\paragraph{Routing.}
``A question about a long recording asks about this event: \{event\}. Which category describes the event best?''
The eight options are: the playback speed suddenly becomes faster, the playback speed suddenly becomes slower, the volume suddenly becomes louder, the volume suddenly becomes quieter, a stretch of silence where the audio is missing, a stretch of noise that makes the speech unintelligible, a particular sound or sound effect that can be heard, and none of these.
The answer is then the option letter with the largest logit.

\paragraph{Answering.}
The reader receives the selected excerpts as separate audio segments, together with the original prompt of the benchmark.
One line per segment states its start and end as minutes and seconds of the original recording.
For occurrence chains there is one 60-second segment.
Native multiple-choice questions use three 20-second windows, with adjacent windows merged.

\section{Further Details for Reproduction}
\label{app:repro}

\paragraph{Baselines.}
The learned retriever and candidate verification are our own baselines.
The retriever presents eight regions of 4 to 8 seconds with their times in seconds.
Candidate verification asks the reader a yes/no question about 16 candidates of the learned scorer.
It answers Q1 with the option that matches its number of verified candidates and the time of its $k$-th verified candidate.
For Q2 and Q3, it hears again at that candidate.
The option rule chooses Q1 without listening.
It takes the earlier option time when $k=1$ and the later one otherwise, together with the smaller of the two option counts.
For Q2 and Q3, the reader then hears the 60-second excerpt centred on the chosen time.

\paragraph{Evaluation Details.}
We rebuild the edited LongAudioSpan recordings with the official preparation script.
On VoiceGiraffe, the time labels of the excerpts are in seconds, since its options give times in seconds.
On LAT-Bench, the scan prompt has no options, since its queries have none.
For grounding, the expanded positive-score region extends from the start of the first included window to the end of the last, as defined in Section~\ref{sec:content}.
BM25 uses $k_1 = 1.5$ and $b = 0.75$.
The calibrators are logistic regressions with an L2 penalty of 1.
The development numbers of the occurrence path use five-fold cross-fitting over recordings.
Intervals come from 2{,}000 bootstrap resamples of recordings.
Paired comparisons count the questions that only one of two systems answers correctly and apply a two-sided sign test over questions or chains.
The full recording with time marks is given as consecutive 30-second audio segments.
Each segment is preceded by a text item with its start time, and one sentence before them explains the format.
One 80\,GB device holds at most about 62 minutes of audio in one input.
The full-recording runs therefore divide the reader's decoder layers between two devices at a layer boundary.
The VoiceGiraffe results use the native 16\,kHz waveforms of the benchmark, and answers without an option letter take the option with the highest log-probability.
On VoiceGiraffe, the scan scores 16 windows per batch, whereas it scores 32 on LongAudioSpan and LAT-Bench.
In BF16, the score of a window varies slightly with the shape of its batch.
With 32 windows per batch, the VoiceGiraffe scan selects at least one different window for 952 of the 1{,}499 questions.
ReListen then reaches 62.6\% instead of 62.0\%, a difference of 0.5 points with a 95\% interval from $-0.7$ to 1.7.
The LongAudioSpan matched-content comparison instead hears slices of the reader's encoding of each whole recording, placed where the excerpt would stand.
On all native questions, ReListen answers 54.7\% from such slices and 55.1\% from waveforms, a difference that is not significant.

\paragraph{Cost.}
All methods run on A100 80\,GB GPUs.
We exclude model loading and one warm-up question, and we synchronize the GPU at the start and end of each stage.
Each text-mediated time includes transcribing the whole recording.
All weights are in BF16, and the scan scores 32 windows per batch.
When the full recording exceeds one GPU, the reader is divided over two GPUs, as in all full-recording runs.
The reader and Whisper run on one local GPU.
The parallel scan, the two-GPU full recording, the vLLM transcription of Speaker-Reasoner and the official LAT-Audio runner run on cluster nodes with the same GPU type.
On LongAudioSpan, a scan makes one reader call per window, 295 on average.
On one A100 80\,GB GPU, the reader scores about 24 windows per second.
Its peak memory stays below 68\,GiB at every duration, including 59\,GiB of weights.
On the 22 timed LongAudioSpan recordings, 61 minutes on average, the scan takes 15.0 seconds and the 60-second answer 0.4 seconds.
The reader writes its transcript of these recordings in 116 seconds on average, and Whisper large-v3 in 24.7 seconds.
The full recording fits one GPU up to 55 minutes on LongAudioSpan and 61.6 minutes on VoiceGiraffe, and it runs out of memory on every longer timed recording.
Its forward pass takes 0.13 seconds per minute of audio at 10 minutes and 0.20 seconds at 62 minutes.
On one GPU, ReListen-Fast is faster than the full recording on 12 of the 13 timed recordings between 30 and 62 minutes and slower on all 7 recordings under 30 minutes.
Beyond 62 minutes, ReListen-Fast on one GPU is faster than the full recording on two GPUs on 10 of the 12 timed recordings.
The two exceptions last over 2.5 hours, of which the full recording hears only the first 80 minutes.
ReListen-Fast halves the scan and lowers its hit rate from 68.5\% to 64.2\% on development recordings (Section~\ref{sec:cost}).
Each GPU of the parallel scan scores whole batches of 32 windows, so short recordings give some GPUs fewer batches or none and gain less from four GPUs.
The three timed questions on which the four-GPU scan is slower than the full recording are on 10-minute recordings, whose batches fill only two of the four GPUs.
On every timed question, the four-GPU scan reproduces the one-GPU scores and selections for both hops.
On one GPU, the cluster nodes take 5 to 8\% longer than the local GPU, so the four-GPU speedups in Section~\ref{sec:cost} understate those within one node.
LAT-Audio runs one of the ten timed LAT-Bench queries, on a 29.7-minute recording, over two GPUs.
On LAT-Bench, the peak memory per GPU is 62.8\,GiB for LAT-Audio and 64.8\,GiB for the scan.

\section{Comparisons with Closely Related Systems}
\label{app:prior-art}

We compare ReListen with the three closest re-listening systems on shared questions and scoring.

\paragraph{LAT-Audio.}
We run the released LAT-Audio checkpoint \citep{shao2026listeningwithtime} with its official runner on the same 1{,}002 queries as ReListen.
Each metric is computed within a language, and the two languages are weighted equally.
The released model reaches a macro average of 46.3 mIoU, with seven failed queries counted as zero.
The authors report 48.6 under their original protocol, which averages over parsed outputs of 1{,}006 queries.

\paragraph{Echo-Style Frozen Adaptation.}
Echo \citep{wu2026echo} learns audio-interleaved reasoning with supervised fine-tuning and reinforcement learning on Qwen2.5-Omni-7B.
Its trained checkpoint is unavailable, so we pair its interface with the frozen Qwen3-Omni reader.
The reader starts from its own whole, timestamped transcript and may request waveform excerpts while it reasons.
We allow eight turns and at most 60 seconds of audio in total with greedy decoding.
A paired control uses the same prompt and transcript but cannot receive audio.
On VoiceGiraffe, the adaptation receives audio for only 22 of 1{,}499 questions, 413 seconds in total.
Most of its requests repeat the placeholder times of the format instruction.
Its 58.1\% accuracy differs from the control's 58.2\% on three questions.
On the 540 test LongAudioSpan multiple-choice questions, it receives audio for 13 questions, 400 seconds in total.
It reaches 58.1\% against 58.3\% for the control, a difference of one question.
Thus, with a frozen reader, the Echo interface rarely returns to the audio, and its accuracy is that of its transcript.

\paragraph{Speaker-Reasoner Transcript Frontend.}
We use the released bilingual Stage 2 checkpoint of Speaker-Reasoner \citep{lin2026speakerreasoner} with its official prompts.
This checkpoint lacks the Stage 3 cache that tracks speakers across chunks.
We therefore transcribe each recording in 45-second chunks with speaker labels local to each chunk.
Chunks that still fail after the official retries remain gaps in the transcript, 212 of 9{,}103 on VoiceGiraffe.
Questions over gaps stay in the denominators.
The frozen reader answers from this transcript with a 60{,}000-token input budget.
On VoiceGiraffe it answers 835 of 1{,}499 questions correctly.
On LongAudioSpan, the transcripts of 21 test recordings exceed the budget and are cut.
The reader answers 286 of the 540 test multiple-choice questions correctly (53.0\%).

\section{Native Questions in Detail}
\label{app:content}

We give the full native-question results behind Section~\ref{sec:content-results}, by option form, recording length and question dimension.
The duration tiers are short (S, below 30 minutes), medium (M, 30--60 minutes) and long (L, above 60 minutes).

\begin{table*}[t]
\centering
\small
\setlength{\tabcolsep}{4.5pt}
\begin{tabular}{llccccccc@{\hskip 14pt}ccc}
\toprule
 & & \multicolumn{7}{c}{Test} & \multicolumn{3}{c}{Development} \\
\cmidrule(lr){3-9}\cmidrule(lr){10-12}
Method & Audio & All & Content & Time & S & M & L & Hit & All & Content & Time \\
\midrule
\grouprow{12}{Text-mediated}
Whisper transcript & none & 58.1 & \textbf{67.5} & 35.4 & 58.2 & 58.7 & \textbf{57.5} & -- & 54.8 & \textbf{65.8} & 30.1 \\
Whisper text + ReListen & 60 s & \textbf{59.8} & 63.1 & \textbf{51.9} & \textbf{61.6} & \textbf{60.8} & 56.9 & 69.2 & \textbf{58.9} & 65.0 & \textbf{45.2} \\
\midrule
\grouprow{12}{Audio input}
No audio & none & 43.5 & 45.3 & 39.2 & 42.4 & 42.3 & 46.0 & -- & 44.8 & 46.0 & 42.2 \\
Uniform & 60 s & 45.0 & 49.0 & 35.4 & 46.9 & 42.9 & 45.4 & 18.7 & 44.1 & 49.7 & 31.3 \\
Learned retrieval & 60 s & 44.6 & 50.3 & 31.0 & 43.5 & 44.4 & 46.0 & 32.6 & 45.7 & 50.8 & 34.3 \\
BM25-selected audio & 60 s & 52.4 & 55.5 & \textbf{44.9} & 53.7 & 53.4 & 50.0 & 45.2 & 48.1 & 50.3 & 43.4 \\
Full recording, time marks & $\leq$80 min & 54.6 & 59.9 & 41.8 & \textbf{62.1} & \textbf{57.1} & 44.3 & \textbf{97.1} & \textbf{58.0} & 62.8 & 47.0 \\
ReListen, more audio & 600 s & \textbf{56.5} & \textbf{64.1} & 38.0 & 61.0 & 55.6 & \textbf{52.9} & 94.7 & 54.8 & \textbf{64.2} & 33.7 \\
ReListen & 60 s & 53.5 & 57.3 & 44.3 & 55.4 & 55.6 & 49.4 & 69.2 & 56.7 & 60.4 & \textbf{48.2} \\
\midrule
\grouprow{12}{Reference}
Annotated evidence & $\leq$120 s & 72.7 & 73.0 & 72.0 & 69.9 & 73.8 & 74.3 & 100.0 & 72.9 & 71.1 & 77.0 \\
\bottomrule
\end{tabular}
\caption{\textbf{Native multiple-choice results on LongAudioSpan.}
We report accuracy (\%) by option type and recording length, and the evidence hit rate (\%) of the audio heard.
Whisper text + ReListen adds the selected minute to the Whisper transcript.
Annotated evidence is scored on the 487 test questions with annotated support ranges.}
\label{tab:native}
\end{table*}

\paragraph{The Full Recording.}
The full recording gives the reader the first 80 minutes of every recording, as the benchmark's authors do.
The audio is one item, or 30-second segments with time marks.
Over all 1{,}080 native questions, the reader answers 54.8\% without time marks, where the benchmark's authors report 51.5\%.
In the three duration tiers, it answers 58.1\%, 56.7\% and 49.7\%.
With time marks, it answers 56.3\%.
On the test half the two yield 54.4\% and 54.6\%, and 58.0\% with time marks on development recordings.
The chain score over all items is 16.1 without time marks, where the authors report 17.6, and 17.7 with them.
With time marks, the full recording and ReListen do not differ significantly on the test half, 54.6\% against 53.5\%.
Their order depends on recording length.
On recordings under 30 minutes, the full recording leads, 62.1\% against 55.4\%.
Beyond an hour, where recordings longer than 80 minutes are cut, ReListen scores higher, 49.4\% against 44.3\%.
Neither difference is significant.
The development half repeats both orders.

\begin{table*}[t]
\centering
\small
\setlength{\tabcolsep}{3.8pt}
\begin{tabular}{lccccccc@{\hskip 12pt}cc}
\toprule
 & & & \multicolumn{5}{c}{Audio input} & \multicolumn{2}{c}{Reference} \\
\cmidrule(lr){4-8}\cmidrule(lr){9-10}
Dimension & $n$ & Time options & No audio & Uniform & ReListen & ReListen 600 s & Full recording & Transcript & Oracle \\
\midrule
\grouprow{10}{Perception}
Recognition & 48 & 0 & 58.3 & 56.3 & 75.0 & \textbf{77.1} & 68.8 & 72.9 & 79.5 \\
Localization & 80 & 78 & \textbf{43.8} & 40.0 & 40.0 & 31.3 & 38.8 & 41.2 & 73.2 \\
Counting & 52 & 0 & 48.1 & 34.6 & 36.5 & 46.2 & \textbf{51.9} & 48.1 & 55.3 \\
\grouprow{10}{Understanding}
Tracking & 81 & 69 & 29.6 & 29.6 & 48.1 & 48.1 & \textbf{51.9} & 34.6 & 70.7 \\
Interpretation & 58 & 1 & 50.0 & 56.9 & 69.0 & \textbf{72.4} & \textbf{72.4} & 69.0 & 71.9 \\
Summarization & 41 & 7 & 51.2 & 48.8 & 58.5 & \textbf{65.9} & 53.7 & 78.0 & 64.7 \\
\grouprow{10}{Reasoning}
Causal & 57 & 0 & 45.6 & 49.1 & 59.6 & \textbf{75.4} & 64.9 & 73.7 & 88.9 \\
Logical & 61 & 3 & 50.8 & \textbf{59.0} & \textbf{59.0} & 54.1 & 54.1 & 68.9 & 72.9 \\
Relational & 62 & 0 & 25.8 & 40.3 & 46.8 & \textbf{56.5} & 45.2 & 59.7 & 71.7 \\
\bottomrule
\end{tabular}
\caption{\textbf{Native accuracy by question dimension.}
We report test accuracy (\%), with $n$ questions per dimension and the number of them whose options are clock times.
The full recording carries time marks, the transcript is from Whisper, and the oracle gives the reader the annotated support ranges.}
\label{tab:dimensions}
\end{table*}

\paragraph{Results by Dimension.}
Table~\ref{tab:dimensions} splits the test questions by dimension.
One selected minute helps most on recognition, tracking, interpretation and relational reasoning, by 17 to 21 points over no audio.
However, it does not help on localization, where the options are clock times.
Nevertheless, the oracle is higher there, since its excerpts are labelled with the times that the options name.
Without time marks, the transcript falls to 34.6\% on tracking, where most options are clock times, against 48.1\% for ReListen.
Hence, the benefit of the selected minute depends on the form of the options.
On counting, the selected minute also falls below no audio, 36.5\% against 48.1\%.
Counting needs every occurrence of an event, which selection by relevance does not guarantee (Section~\ref{sec:occurrence}).
Ten minutes raise counting to 46.2\%.
The full recording reaches 51.9\%.
The occurrence extension addresses this limit for planted events, but we do not apply it to native counting questions.

\paragraph{Transcript Systems.}
Whisper large-v3 transcribes each recording in consecutive 30-second chunks with greedy decoding and the language of the recording forced.
Chunks whose text compresses by a factor above 2.4 are treated as repetition loops and dropped, 0.6\% of all chunks.
The transcript given to the reader joins the chunk texts in order, without time marks.
The longest resulting input has 33{,}701 tokens, within the reader's context, so no transcript is truncated.
For BM25, the query is the question with its options, and the ranking is computed within the recording.
Tokens are lower-cased words for English, and character unigrams and bigrams for Mandarin after conversion to simplified characters.

\section{Questions about the Whole Recording}
\label{app:global}

\begin{table}[h]
\centering
\small
\setlength{\tabcolsep}{3.5pt}
\begin{tabular}{lccccc}
\toprule
 & \multicolumn{2}{c}{Half} & \multicolumn{3}{c}{Duration tier} \\
\cmidrule(lr){2-3}\cmidrule(l){4-6}
Input & Test & Dev & S & M & L \\
 & (61) & (72) & (61) & (39) & (33) \\
\midrule
No audio & 49.2 & 55.6 & 49.2 & 61.5 & 48.5 \\
ReListen, 60 s & 55.7 & 62.5 & 62.3 & 56.4 & 57.6 \\
ReListen, 180 s & 67.2 & 68.1 & 68.9 & 64.1 & 69.7 \\
ReListen, 300 s & 68.9 & 69.4 & 72.1 & 64.1 & 69.7 \\
ReListen, 600 s & 72.1 & 75.0 & \textbf{80.3} & 61.5 & \textbf{75.8} \\
Full recording & \textbf{73.8} & 77.8 & \textbf{80.3} & 71.8 & 72.7 \\
Whisper transcript & 68.9 & \textbf{79.2} & 75.4 & 71.8 & \textbf{75.8} \\
Reader transcript & 72.1 & 75.0 & 75.4 & \textbf{76.9} & 66.7 \\
\midrule
Annotated evidence & 72.1 & 70.8 & 68.9 & 74.4 & 72.7 \\
\bottomrule
\end{tabular}
\caption{\textbf{Questions about the whole recording.}
We report native accuracy (\%) on LongAudioSpan questions whose annotated evidence extends over at least half of the recording.
Parentheses give the number of questions, and the duration tiers pool both halves.
The full recording carries time marks, and the transcripts carry none.}
\label{tab:global}
\end{table}

We give the details behind Section~\ref{sec:global-results}.
The \emph{extent} of a question's evidence is the time from the start of its first annotated support range to the end of its last, as a share of the recording.
On LongAudioSpan, 61 test and 72 development questions have an extent of at least one half.
Their annotated evidence lasts 170 seconds at the median.
However, 16.5\% of them annotate more than ten minutes of evidence, against 3.7\% of all annotated native questions.
Table~\ref{tab:global} gives their accuracy by budget and by duration tier.

On each half, the selected minute wins 4 of these questions against the full recording and loses 15, a significant difference.
In contrast, ten minutes from the same scan win 3 and lose 4 on the test half, a difference that is not significant.
On the development half they win 4 and lose 6, which is also not significant.
The effect of ten minutes depends on the duration tier.
On short recordings, ten minutes are at least a third of the audio and match the full recording.
On medium recordings, they still trail it, 61.5\% against 71.8\%.
On long recordings, where ten minutes are at most a sixth of the audio, they reach 75.8\% against 72.7\%.
On the other 479 test questions, the selected minute and the full recording perform alike, 53.2\% against 52.2\%.
On VoiceGiraffe, 31 questions have an extent of at least one half.
ReListen answers 54.8\% of them, against 35.5\% for the full recording with time marks.

\section{VoiceGiraffe and LAT-Bench in Detail}
\label{app:voicegiraffe}

We give the per-family VoiceGiraffe results and the LAT-Bench protocol behind Sections~\ref{sec:content-results} and~\ref{sec:locate-results}.
We exclude one malformed Paralinguistic question with an extra empty option.
The sound families are Audio Event (200) and Paralinguistic (349), and the content families are Semantic Content, Temporal Localization, Causal Alignment and Event Tracking.
The options of Temporal Localization are time ranges.
Evidence time ranges exist for 1{,}498 questions, 548 of them sound questions.

\begin{table*}[t]
\centering
\small
\setlength{\tabcolsep}{4.5pt}
\begin{tabular}{lccccccccc}
\toprule
Method & All & Aud. Ev. & Paraling. & Semantic & Temp. Loc. & Causal & Ev. Track. & En & Zh \\
\midrule
\grouprow{10}{Text-mediated}
Transcript by the reader & 52.8 & 46.0 & 53.6 & 85.5 & 38.4 & 46.0 & \textbf{52.4} & 51.1 & 54.6 \\
Transcript + ReListen, 60 s & \textbf{63.0} & \textbf{55.5} & \textbf{61.0} & \textbf{86.0} & \textbf{77.6} & \textbf{49.6} & \textbf{52.4} & \textbf{60.5} & \textbf{65.6} \\
\midrule
\grouprow{10}{Audio input}
No audio & 34.5 & 35.5 & 38.4 & 32.0 & 34.0 & 38.0 & 27.2 & 33.9 & 35.1 \\
Uniform, 60 s & 40.7 & 46.0 & 49.9 & 38.5 & 36.4 & 45.2 & 25.2 & 40.9 & 40.5 \\
Full recording & 49.6 & 49.0 & 53.6 & 70.0 & 33.2 & \textbf{48.4} & 45.6 & 46.9 & 52.2 \\
Full recording, time marks & 51.4 & 49.5 & 52.4 & 67.0 & 48.0 & 46.8 & 47.2 & 48.7 & 54.2 \\
ReListen, 60 s & \textbf{62.0} & \textbf{54.5} & \textbf{61.0} & \textbf{87.5} & \textbf{78.0} & 47.2 & \textbf{48.0} & \textbf{59.7} & \textbf{64.4} \\
\midrule
\grouprow{10}{Reference}
Annotated evidence & 74.5 & 70.4 & 66.5 & 98.0 & 98.4 & 69.2 & 51.6 & 74.6 & 74.4 \\
\midrule
\grouprow{10}{Evidence hit rate}
Uniform, 60 s & 11.1 & 6.0 & 3.4 & 6.5 & 5.2 & 17.2 & 29.6 & 9.9 & 12.4 \\
BM25, 60 s & 58.8 & 41.7 & 47.9 & 63.0 & 66.0 & 68.4 & 67.6 & 57.0 & 60.6 \\
Dense retrieval, 60 s & 52.1 & 30.7 & 41.5 & 60.0 & 62.0 & 56.8 & 62.8 & 51.3 & 52.9 \\
ReListen scan, 60 s & \textbf{88.5} & \textbf{72.4} & \textbf{78.8} & \textbf{98.0} & \textbf{98.0} & \textbf{93.6} & \textbf{92.4} & \textbf{89.7} & \textbf{87.2} \\
\bottomrule
\end{tabular}
\caption{\textbf{VoiceGiraffe results by question family and language.}
We report accuracy (\%) on all 1{,}499 questions and the evidence hit rate (\%) on the 1{,}498 with annotated evidence.
Transcript + ReListen adds the selected minute to the reader's transcript.
BM25 and dense retrieval rank chunks of the same transcript.}
\label{tab:voicegiraffe}
\end{table*}

Table~\ref{tab:voicegiraffe} gives the native-waveform VoiceGiraffe results by question family and language.
We use 1{,}499 of the 1{,}500 questions, excluding one Paralinguistic item with a fifth, empty option.
The Paralinguistic questions ask about the age, gender, emotion, pitch and timbre of a voice.
For them, the benchmark's authors report 50.6\% for the same reader when it reads captions of 30-second windows.
From native VoiceGiraffe waveforms, the full recording answers Temporal Localization questions at 33.2\% without time marks and 48.0\% with them.
Without time marks, the reader's transcript answers 38.4\% of these questions.
The selected minute with its scan-derived time labels reaches 77.6\%, with or without the transcript.
The recordings last 35 to 79 minutes, all within the 80 minutes that the full recording hears.
ReListen exceeds the full recording with time marks, 62.0\% against 51.4\%.

On LAT-Bench, the uniform baseline is one 20-second window at the centre of the recording.
The transcript baseline is the best 30-second chunk of a Whisper transcript under BM25.
The expanded positive-score region reaches a mean IoU of 47.2 on the Chinese queries and 48.2 on the English ones.
We also give the reader the whole recording and ask for the span, as the benchmark does.
We parse the first two times of its answer as the span.
In a second variant, the start time of every 30-second segment is given as text.
Thus, on both benchmarks, the scan's labelled windows exceed the full recording on temporal questions, even when the full recording carries time marks (Table~\ref{tab:grounding}).

\begin{table}[h]
\centering
\small
\setlength{\tabcolsep}{2.4pt}
\begin{tabular}{lcccc}
\toprule
Variant & mIoU & R@0.3 & R@0.5 & R@0.7 \\
\midrule
Best window only & 41.8 & 67.3 & 35.7 & 13.8 \\
Qwen2.5-Omni-7B reader & 46.2 & 75.6 & 45.2 & 17.3 \\
\midrule
ReListen & \textbf{47.7} & \textbf{77.2} & \textbf{50.0} & \textbf{19.2} \\
\bottomrule
\end{tabular}
\caption{\textbf{Grounding ablations on LAT-Bench.}
The first row returns only the highest-scoring window instead of the expanded region.
The second row replaces the reader, Qwen3-Omni-30B-A3B, with Qwen2.5-Omni-7B.
We report mean IoU and recall at IoU thresholds 0.3, 0.5 and 0.7 (\%).}
\label{tab:grounding-ablation}
\end{table}

Table~\ref{tab:grounding-ablation} gives two ablations of the grounding rule.
The expanded region raises the mIoU of the best window by 5.9 points.
With the smaller Qwen2.5-Omni-7B reader, the scan still reaches 46.2 mIoU.

\section{Robustness of the Jev-Style Scan}
\label{app:checks}

We test whether the scan depends on the reader, and whether other locators can replace it.

\paragraph{A Second Reader.}
We repeated the multiple-choice experiment with Qwen2.5-Omni-7B \citep{xu2025qwen25omni} as scanner and answerer.
It uses the configuration of the main reader.
This reader takes at most five minutes of audio per input, so the full recording is not an option for it.
Its scan places a support range in the selected 60 seconds for 61.2\% of the annotated test questions.
It answers 51.9\% of the test questions, against 43.7\% from uniform windows and 39.6\% without audio.
On development recordings the three accuracies are 48.9\%, 43.0\% and 39.3\%.
On the test half, the gain over uniform windows is 6.3 points with content options.
With time options, it is 12.7 points.
On LAT-Bench, the scan of this reader uses the same prompt and windows.
Its best window overlaps the annotated span for 94.1\% of the queries and reaches a mean IoU of 41.6.
The expanded positive-score region reaches 46.2, with recalls of 75.6\%, 45.2\% and 17.3\% at the three thresholds.
By comparison, the main reader reaches 41.8 and 47.7.

\paragraph{A Contrastive Audio-Text Model as Scanner.}
We replaced the reader's scan by CLAP \citep{wu2023laionclap}, a zero-shot audio-text model, on the 92 English test recordings.
On native questions, CLAP places a support range in the selected 60 seconds for 33.9\% of the English test questions.
The reader's scan does so for 68.5\%.
On occurrence chains, however, the two scanners perform alike.
Time accuracy is 50.0\% with either evidence alone.
When each is fused with the detectors, it is 77.2\% with CLAP against 80.4\% with the reader's scan.
Planted sounds and masks are therefore distinct enough for a small contrastive model.
Questions about the content of natural recordings, however, need the reader's understanding of the question.

\paragraph{A Dense Text Retriever.}
BGE-M3 \citep{chen2024bgem3}, a multilingual dense retriever, ranks the same 30-second transcript chunks as BM25 for the same query.
Its two best chunks overlap a support range for 47.8\% of the annotated test questions and 47.5\% of the development questions.
BM25 reaches 45.2\% and 46.5\%, and the scan 69.2\% and 68.5\%.
On native-waveform VoiceGiraffe, the same query and top-two-chunk policy uses the reader's transcript from all 123 recordings.
BM25 hits annotated evidence on 881 of 1,498 questions (58.8\%) and BGE-M3 on 780 (52.1\%), compared with 1,325 (88.5\%) for the audio scan.
On the sound families their hit rates are 45.6\% and 37.6\%, and on the content families 66.4\% and 60.4\%.
BGE-M3 uses normalized CLS embeddings, cosine ranking and a 1{,}024-token encoding limit.
Across these checks, no alternative locator matches the reader's audio scan, in line with Section~\ref{sec:locate-results}.

\section{Scores along Recordings}
\label{app:evidence}

\begin{figure*}[h]
\centering
\definecolor{rlReader}{HTML}{2A78D6}
\definecolor{rlDetector}{HTML}{EB6834}
\definecolor{rlGrid}{HTML}{E4E3DF}
\definecolor{rlMuted}{HTML}{52514E}
\begingroup\figfont%
\begin{tikzpicture}
\begin{groupplot}[
  group style={group size=2 by 2, horizontal sep=0.9cm, vertical sep=0.22cm, x descriptions at=edge bottom},
  width=0.405\textwidth, height=1.55cm, scale only axis=true,
  axis lines=left, axis line style={line width=0.4pt, rlMuted}, tick style={line width=0.4pt, rlMuted},
  tick label style={font=\scriptsize}, label style={font=\scriptsize}, title style={font=\scriptsize, yshift=5pt},
  ymajorgrids, grid style={line width=0.4pt, rlGrid}, clip=false, enlarge x limits=false,
  every axis plot/.append style={line width=0.7pt, line join=round}]
\nextgroupplot[title={Inserted sound: ``hands clapping in applause'', $k=2$}, xmin=0, xmax=57.56, ymin=-9, ymax=9, ytick={-6,0,6},
  ylabel={Reader $\ell$}]
\foreach \x in {15.44,34.09,53.21} {\edef\temp{\noexpand\draw[rlGrid, line width=2pt] (axis cs:\x,-9) -- (axis cs:\x,9);}\temp}
\addplot[rlReader] table[x=t, y=lo] {figures/evidence_reader_a.dat};
\draw[rlMuted, line width=0.4pt] (axis cs:0,0) -- (axis cs:57.56,0);
\addplot[only marks, mark=triangle*, mark options={rotate=180, fill=black, draw=black}, mark size=1.6pt] coordinates {(15.44,10.4) (34.09,10.4) (53.21,10.4)};
\node[font=\scriptsize, anchor=east] at (axis cs:14.2,10.4) {true instances};
\addplot[only marks, mark=o, mark options={draw=black, line width=0.5pt}, mark size=3.2pt] coordinates {(34.09,10.4)};
\node[font=\scriptsize, anchor=west] at (axis cs:35.8,10.4) {selected $t^\star$};
\nextgroupplot[title={Volume decrease, $k=1$}, xmin=0, xmax=149.22, ymin=-9, ymax=9, ytick={-6,0,6}]
\foreach \x in {31.14,70.08,126.29} {\edef\temp{\noexpand\draw[rlGrid, line width=2pt] (axis cs:\x,-9) -- (axis cs:\x,9);}\temp}
\addplot[rlReader] table[x=t, y=lo] {figures/evidence_reader_b.dat};
\draw[rlMuted, line width=0.4pt] (axis cs:0,0) -- (axis cs:149.22,0);
\addplot[only marks, mark=triangle*, mark options={rotate=180, fill=black, draw=black}, mark size=1.6pt] coordinates {(31.14,10.4) (70.08,10.4) (126.29,10.4)};
\addplot[only marks, mark=o, mark options={draw=black, line width=0.5pt}, mark size=3.2pt] coordinates {(31.15,10.4)};
\node[font=\scriptsize, anchor=west] at (axis cs:35.5,10.4) {selected $t^\star$};
\nextgroupplot[xmin=0, xmax=57.56, ymin=0, ymax=1.08, ytick={0,0.5,1}, xlabel={Time (minutes)}, ylabel={Detector $p$}]
\foreach \x in {15.44,34.09,53.21} {\edef\temp{\noexpand\draw[rlGrid, line width=2pt] (axis cs:\x,0) -- (axis cs:\x,1.08);}\temp}
\addplot[rlDetector] table[x=t, y=p] {figures/evidence_detector_a.dat};
\nextgroupplot[xmin=0, xmax=149.22, ymin=0, ymax=1.08, ytick={0,0.5,1}, xlabel={Time (minutes)}]
\foreach \x in {31.14,70.08,126.29} {\edef\temp{\noexpand\draw[rlGrid, line width=2pt] (axis cs:\x,0) -- (axis cs:\x,1.08);}\temp}
\addplot[rlDetector] table[x=t, y=p] {figures/evidence_detector_b.dat};
\end{groupplot}
\end{tikzpicture}%
\endgroup
\caption{\textbf{Reader and detector evidence on planted events.}
On two development recordings, the reader's presence score rises at each inserted sound but stays flat at each volume edit, whereas the acoustic detectors respond to both.
Grey bars and triangles mark the true occurrences, and the circle marks the occurrence that ReListen selected.}
\label{fig:evidence}
\end{figure*}
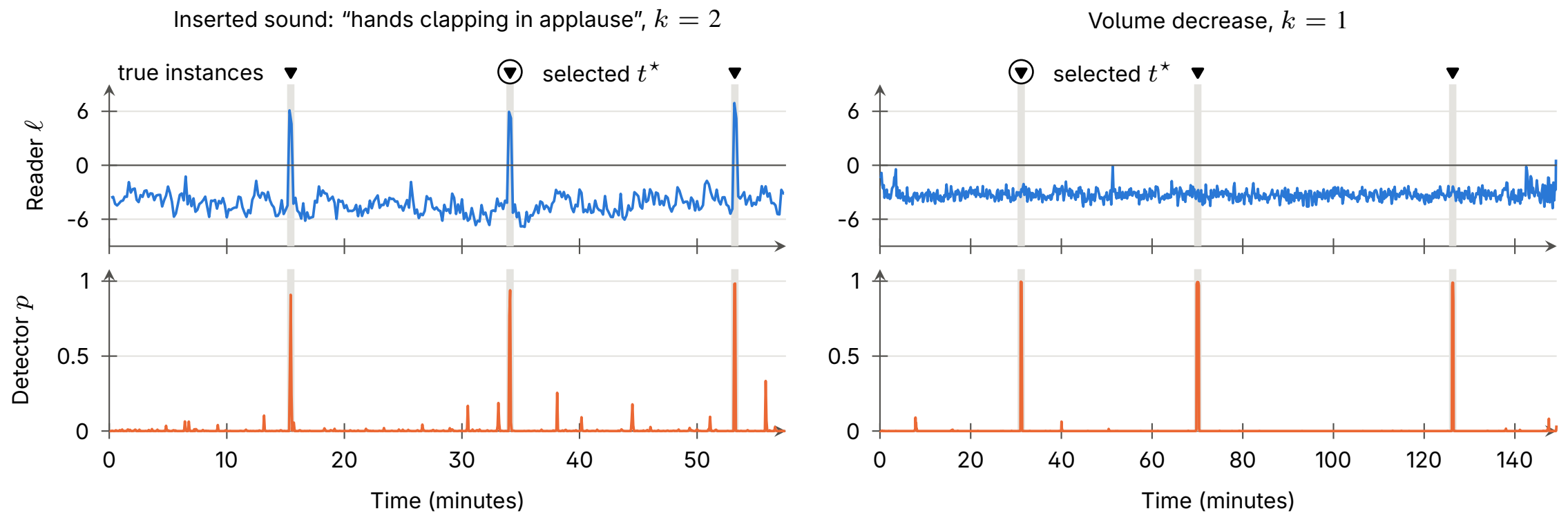

Figure~\ref{fig:profile} shows the scan of a native question from a development recording, and Figure~\ref{fig:evidence} shows the reader's presence scores and the detector outputs for two chains.
On the test question of Figure~\ref{fig:overview}, the reader also answers 4 without audio and from the full recording, and 3 from the Whisper transcript.

\begin{figure}[h]
\centering
\begingroup\figfont%
\begin{tikzpicture}
\begin{axis}[width=\columnwidth, height=3.7cm, xmin=0, xmax=87, ymin=-5, ymax=9.5,
  xlabel={Time in the recording (min)}, ylabel={Score $\ell_i$}, tick label style={font=\scriptsize}, label style={font=\footnotesize},
  ytick={-4,0,4,8}, xtick={0,20,40,60,80}, ymajorgrids, grid style={black!10}, clip=true]
\fill[black!25] (axis cs:57.68,-5) rectangle (axis cs:58.20,9.5);
\addplot[blue!65!black, line width=0.45pt] table[x=minute, y=score] {figures/profile_native.dat};
\addplot[only marks, mark=o, mark size=2.3pt, line width=0.8pt, orange!85!black] table[x=minute, y=score] {figures/profile_native_sel.dat};
\node[font=\scriptsize, anchor=south east, text=black!70] at (axis cs:57.4,6.6) {annotated evidence};
\end{axis}
\end{tikzpicture}%
\endgroup
\caption{\textbf{Scan scores along a recording.}
The scan of an 87-minute development recording for the question ``According to the lecture, what is the direct mechanical cause of the lens flattening when shifting focus to a distant object?''
Circles mark the three windows that ReListen selects, and the grey band marks the support range that the benchmark annotates for the question.}
\label{fig:profile}
\end{figure}
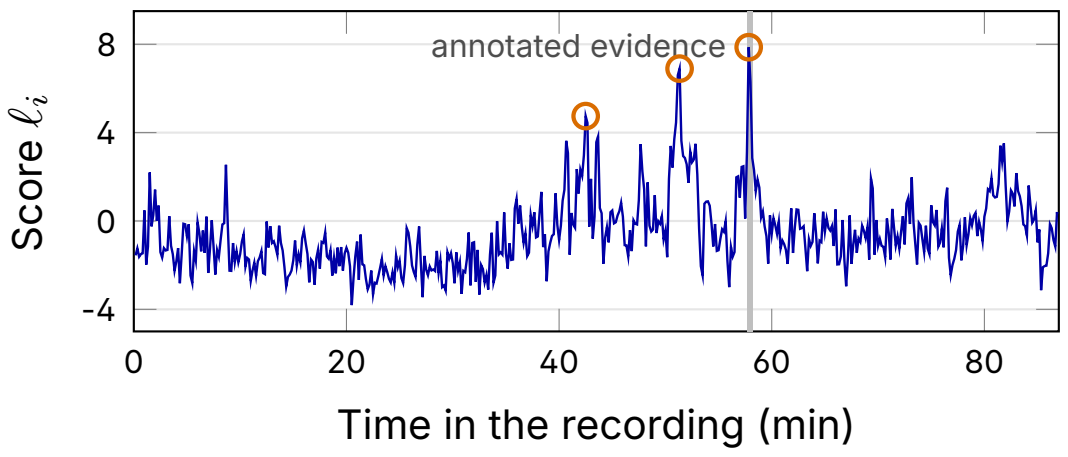

\section{Reading and Listening}
\label{app:reading}

We detail how reading and listening compare when the reader receives its own transcript, the same excerpts as text, or the annotated evidence.

\paragraph{The Reader as Its Own Recognizer.}
The frozen reader transcribed every native recording in consecutive 30-second excerpts, one generation per excerpt and independent of any question.
It then read its own transcript as text, with the prompt and the format of the Whisper transcript.
It answers 59.6\% of the test multiple-choice questions and 57.0\% of the development questions.
The Whisper transcript yields 58.1\% and 54.8\%, a difference that is not significant.
The two transcripts are level on questions with content options, 67.2\% against 66.7\% over both halves.
However, the reader's transcript is ahead on questions with time options, 37.7\% against 32.7\%.
The lead of a transcript on native questions therefore needs no second system.

\paragraph{Reading and Listening at Matched Content.}
\label{app:matched}
We use two selections, the frozen scan's three windows and the annotated evidence.
For each question and selection, the reader transcribes exactly the selected excerpts.
It then answers from that text without audio.
An excerpt is transcribed in equal pieces of at most 30 seconds, and the text uses the same time label as the audio input.
A second text input contains notes with the words and a description of voices and other sounds.
The LongAudioSpan control uses the encoder once on each whole recording and selects slices from its outputs.
The main VoiceGiraffe control instead crops the native waveform before encoding, consistently across the audio, transcript and notes inputs.

On LongAudioSpan questions without time options, listening and reading the scan's windows yield 60.5\% and 59.4\% on the test half.
On the development half, they yield 59.9\% and 59.1\%.
From annotated evidence the corresponding values are 70.9\% and 70.1\%, and 71.7\% and 71.7\%.
Neither difference on the test half is significant.
With time options, text is ahead from the scan's windows, 51.9\% against 42.0\% over both halves.
It is also ahead from annotated evidence, 79.0\% against 71.5\%.
The scan-window difference is 5.7 points on the test half and 13.8 on the development half.

On native-waveform VoiceGiraffe, audio and matched transcripts yield 62.0\% and 59.6\% from the scan's excerpts.
From annotated evidence, they yield 74.5\% and 66.1\%.
On content questions the audio-minus-transcript differences are $-1.2$ points for the scan and 2.0 points for annotated evidence.
On sound questions the differences are 8.6 and 19.5 points, and both are significant.
For annotated Audio Event questions, audio reaches 70.4\% against 39.7\% for the transcript.
For Paralinguistic questions, it reaches 66.5\% against 53.3\%.
Notes reach 57.7\% on sound questions from the scan's excerpts and 58.2\% from annotated evidence, compared with audio at 58.7\% and 67.9\%.
Over all questions, scan notes reach 62.0\%, almost identical to scan audio.
These results distinguish losses in ordinary speech transcripts from the information that richer text descriptions can preserve.

With Whisper text on LongAudioSpan, the two best BM25 chunks yield 51.5\% as text and 52.4\% as audio on the test half.
Whisper chunks overlapping the scan's windows provide a superset of the selected audio content.
They yield 57.6\% against 53.5\% for audio on the test half.

\paragraph{The Gap to Annotated Evidence by Scan Hit.}
On the test half of LongAudioSpan, the questions that the scan misses account for 9.9 of the 19.7 points between ReListen and the annotated evidence.
On the development half, they account for 8.9 of 16.3.
The questions that the scan hits account for 9.9 and 7.5.
The misses thus account for about half of the gap.
However, more audio does not close the gap on the hits.
Consider the 484 questions without time options whose support range the selected minute overlaps.
On them, the reader answers 64.3\%, 63.0\% and 64.7\% from 60, 180 and 600 seconds over both halves (62.0\%, 60.7\% and 63.2\% on the test half).
These inputs hold 38\%, 55\% and 74\% of the annotated seconds, and the evidence alone yields 72.9\%.
On the 203 such questions that the minute misses, however, the three budgets yield 45.3\%, 52.2\% and 59.1\%.
Here, ten minutes win 40 questions against the minute and lose 12.
The evidence yields 70.0\% on them.
Over all questions without time options, ten minutes yield 64.2\% over both halves, against 67.2\% for the reader's transcript and 65.1\% for ten minutes of BM25 text.
On these questions, coverage therefore closes most of the transcript's lead.
By comparison, the reader's transcript leads the selected minute on the test half by 10.0 points on the 382 questions without time options.
On the 158 questions with time options, the minute leads by 3.2 points.

\begin{figure}[h]
\centering
\begingroup\figfont%
\begin{tikzpicture}
\begin{axis}[vizaxis, scale only axis, width=6.6cm, height=3.2cm, xmin=-0.35, xmax=5.35, ymin=35, ymax=80,
  xtick={0,...,5}, xticklabels={0,{0--25},{25--50},{50--75},{75--\textless100},100}, ytick={40,50,60,70,80},
  xlabel={Annotated support seconds in the input (\%)}, ylabel={Accuracy (\%)}]
\addplot[vizblue, vizline, error bars/.cd, y dir=both, y explicit,
  error bar style={vizblue!50, line width=0.5pt}, error mark options={vizblue!50, line width=0.5pt, mark size=1.2pt, rotate=90}]
  table[x=bin, y=acc, y error plus expr=\thisrow{hi}-\thisrow{acc}, y error minus expr=\thisrow{acc}-\thisrow{lo}] {figures/coverage.dat};
\end{axis}
\end{tikzpicture}%
\endgroup
\caption{\textbf{Accuracy against the evidence heard.}
Native multiple-choice questions with content options on both halves of LongAudioSpan, with the scan at 60, 180 and 600 seconds pooled and grouped by the share of the annotated support seconds that the selected audio holds.}
\label{fig:coverage}
\end{figure}
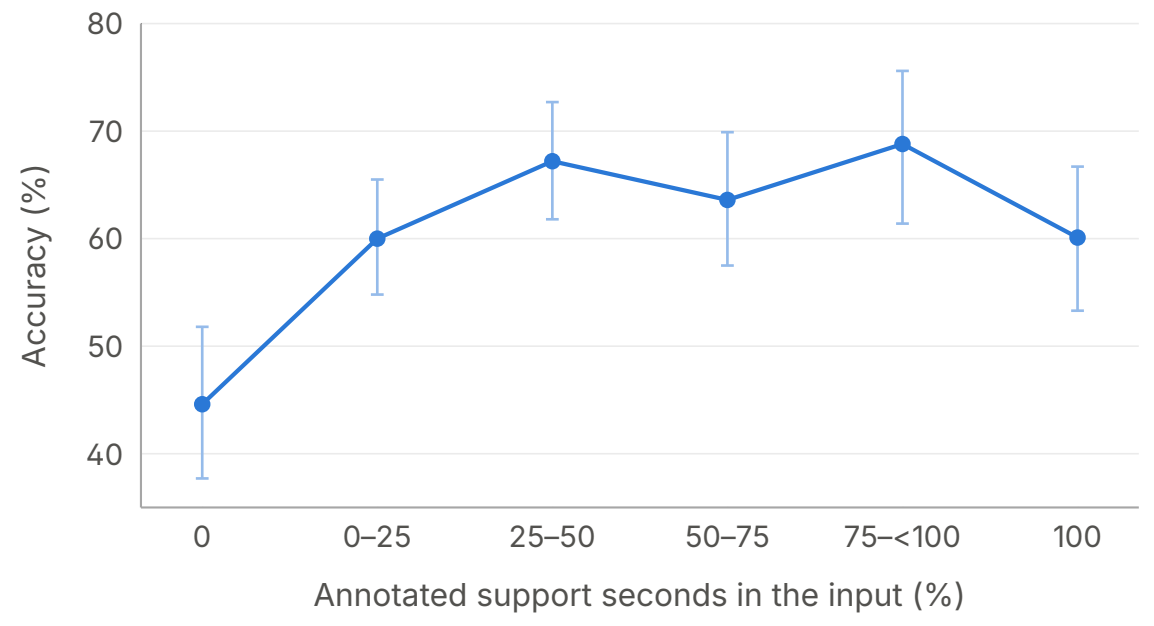

Figure~\ref{fig:coverage} relates accuracy to the share of the annotated support seconds that the selected audio holds.
Without any annotated second the reader answers 44.6\% of these questions, with up to a quarter of them 60.0\%, and with more 63.6\% to 68.8\%.
Fully covered ranges yield only 60.1\%.
They are short, 13 seconds at the median against 78 seconds for partly covered ones, and come from a different mix of question types.
Hence, the largest step is from hearing none of the evidence to hearing part of it.
More of the evidence adds little.
The same holds within a question.
From 60 to 600 seconds, among the questions that the minute misses and ten minutes cover by at least half, 26 become correct and 6 become wrong.
However, among the questions that the minute already hits and whose coverage rises by at least half, 15 become correct and 12 become wrong.
Therefore, the remaining gap on the hits does not come from too little evidence in the input.
It points instead to the reader's use of the evidence it hears.

\section{Occurrence Extension in Detail}
\label{app:occurrence}

We specify the candidates, the detectors, the calibration and the posterior of the occurrence extension in Section~\ref{sec:occurrence}.

\subsection{Candidates}
The candidates are the highest-scoring windows of the scan together with the regions flagged by two small acoustic detectors.
The detectors are convolutional networks with under one million parameters each that read the waveform.
The reader's scores separate inserted sounds and masks from other windows, with areas under the ROC curve of 0.93 to 0.98 on development recordings.
However, they carry little information about changes of speed (0.79) and volume (0.57), so we add the detectors.
We train the detectors on unrelated recordings augmented with the benchmark's public edit operators (Appendix~\ref{app:detector}).
For each of four event groups, we fit a logistic model on development recordings to predict whether a candidate matches an annotated edit.
This calibration combines reader and detector features into a presence probability $p_j$ (Appendix~\ref{app:candidates}).

\subsection{Acoustic Detectors}
\label{app:detector}

\paragraph{Inputs and Architecture.}
The speed and volume detector has 0.63M parameters.
It reads eight band energies, the broadband energy and an onset-strength channel, and predicts volume increase, volume decrease, speed-up and slow-down.
The content-edit detector has 0.71M parameters and reads 40 log-mel bands.
It is trained on all seven edit classes and supplies the probabilities of inserted sound, silence mask and noise mask.
Each is used as an ensemble of three seeds, 4.0M parameters in total.
Four strided convolutions with kernel sizes 7, 5, 5 and 4 reduce 100 frames per second to one.
The intermediate channels are 64, 96 and 128 for the 40-band input and 48, 64 and 96 for the 10-channel input.
Eight residual blocks of width 128 follow.
Each has a dilated convolution of kernel size 3, batch normalization and a point-wise convolution, with dilations 1, 2, 4, 8, 16, 32, 64 and 1 across the blocks.
A point-wise convolution then gives one logit per class and second.

\paragraph{Training Data.}
The detectors' classes are the benchmark's edit types.
Their training edits are produced with the benchmark's public preparation code on unrelated recordings.
These are meetings from AMI \citep{carletta2006ami} and AISHELL-4 \citep{fu2021aishell4}, read speech from LibriSpeech \citep{panayotov2015librispeech}, the 30 podcast episodes in the full Rev.AI collection\footnote{The \texttt{full} configuration of \url{https://huggingface.co/datasets/distil-whisper/rev16}, whose 16-episode subset is the Rev16 evaluation set of \citet{radford2023whisper}.} and talks from TED-LIUM \citep{hernandez2018tedlium3}.
Each training clip is a crop of 8 to 15 minutes with zero to four edits whose parameters are drawn from broad ranges.
These are gains of 5 to 14\,dB, tempo factors of 0.33 to 0.67 and 1.5 to 3, inserted sounds at 3 to 14\,dB above the local level, and masks of 5 to 35 seconds.
Inserted sounds come from ESC-50 \citep{piczak2015esc} and from procedurally generated tones, chirps and pulse trains.

Two choices make the detectors transfer to the benchmark recordings.
The first is background diversification: dynamic-range compression, noise gating to digital silence, additive noise, low-pass filtering, silence removal and peak normalization.
Digital silence occurs in many benchmark recordings and in almost no meeting recording.
The second is negative examples that resemble edits: a segment replaced by another recording at a different level, and a level change with slow ramps.
Level features are expressed relative to the 90th percentile of the clip and clamped 70\,dB below it.
This keeps the operating point fixed under a global gain or digital silence.
Inference runs on 400-second crops with 200-second hop, matching the training length.

\paragraph{Training.}
The loss is binary cross-entropy with a positive weight of 3, summed over all valid second-class pairs of a batch and divided by their number.
Seconds that an edit covers only partly are excluded.
We train with AdamW, a one-cycle schedule with peak learning rate $2\times10^{-3}$, weight decay 0.01, batches of 32 crops of 400 seconds and gradient clipping at 1.
The speed and volume detector trains for 4{,}000 steps on 5{,}000 clips (927 hours of rendered audio).
The content-edit detector trains for 6{,}000 steps on 6{,}000 clips (1{,}121 hours).
The clips are repeated crops of the source recordings.
A random global offset is added to the level channels of each crop.
Checkpoints are selected by the loss on synthetic clips built from held-out sessions, speakers and sound effects.

\paragraph{Synthetic Edits.}
Volume edits multiply a region of 6 to 35 seconds by a gain of 5 to 14\,dB in either direction and clip at full scale.
Speed edits apply the \texttt{atempo} filter of FFmpeg with a factor of 1.5 to 3 or 0.33 to 0.67 to a region whose output lasts 3 to 70 seconds.
Inserted sounds last 0.8 to 9 seconds and are mixed 3 to 14\,dB above the level of the region they cover.
Silence masks set 5 to 35 seconds to zero.
Noise masks replace 5 to 35 seconds by a copy that is low-passed at 200 to 600\,Hz and attenuated by 3 to 12\,dB.
Pink, white or brown noise is added, with fades of 0.2 to 1 second.
Sixty percent of the clips use one edit type for all regions and one repeated sound effect, as the benchmark does, whereas the rest mix types.
Forty-five percent of the backgrounds are podcasts and talks, twenty percent read speech and the rest meetings.

\subsection{Candidate Features and Calibration}
\label{app:candidates}

The reader classifies the event phrase, without audio, into an edit class, which selects the detector output to read.
Candidates are then the union of two sets.
The first is the 24 windows with the highest scores $\ell_i$ whose centres are at least 30 seconds apart.
The second is the regions where the routed detector probability exceeds 0.2.
A reader peak within 15 seconds of a detector region is merged into it.
However, for speed and volume events the candidates are the detector regions alone, since the reader's scores carry little information about them.
A direct question to the reader about loudness does no better, with AUCs of 0.56 to 0.66 on development recordings.
Each candidate has a feature vector with reader features and detector features.
The reader features are the score $\ell$ of the best window covering the candidate, the larger score of its two neighbours, and the score relative to the 90th percentile and to the maximum of the recording.
The detector features are the mean probability in the region and the region length capped at 8 seconds.
Volume candidates add the smaller of the two level steps at the region boundaries.
The logistic models are fitted on the development recordings, with cross-fitting for the development numbers.
Appendix~\ref{app:ablation} measures how many annotated recordings they need.

\subsection{Occurrence Posterior}
Let the candidates be sorted by time, $t_1 < \dots < t_m$, with presence probabilities $p_1, \dots, p_m$ that we treat as independent.
For candidate $j$ to be the $k$-th occurrence, it must be real and exactly $k-1$ earlier candidates must be real.
Thus its probability is
\begin{equation}
\pi_j^{(k)} = p_j \cdot \Pr\Big[\textstyle\sum_{i<j} z_i = k-1\Big],
\end{equation}
with $z_i \sim \mathrm{Bernoulli}(p_i)$.
Here the second factor is a Poisson-binomial probability \citep{hong2013poissonbinomial}.
One pass of dynamic programming computes it for all $j$ and also gives the distribution of the count $N = \sum_j z_j$.
The extension then selects $t^\star$ as the time of the candidate with the largest $\pi_j^{(k)}$.
By comparison, verification pipelines keep the candidates judged present and take the $k$-th in time.
One weak false alarm or miss before the target then shifts the answer.
For Q2 and Q3, the reader hears one 60-second window centred on $t^\star$ and answers the original question.
For Q1, each option names a count $n$ and a time $t$.
ReListen scores an option by $\Pr[N = n] \cdot \sum_{j: |t_j - t| \leq 45\,\mathrm{s}} \pi_j^{(k)}$.

\section{Occurrence Chains: Both Halves and Edit Types}
\label{app:chainfull}

\begin{table*}[t]
\centering
\small
\begin{tabular}{lcccc@{\hskip 18pt}cccc}
\toprule
 & \multicolumn{4}{c}{Test} & \multicolumn{4}{c}{Development} \\
\cmidrule(lr){2-5}\cmidrule(lr){6-9}
Method & Q1 & Q2 & Q3 & Chain & Q1 & Q2 & Q3 & Chain \\
\midrule
\grouprow{9}{Locating without the Q1 options}
No audio & 16.1 & 39.4 & 29.4 & 8.3 & 25.6 & 32.2 & 28.3 & 13.7 \\
Uniform 60 s & 22.8 & 28.9 & 34.4 & 10.0 & 33.3 & 28.9 & 27.8 & 15.0 \\
Learned retrieval & 18.3 & 39.4 & 32.8 & 9.6 & 27.2 & 42.8 & 27.2 & 13.5 \\
Whisper transcript & 22.8 & 29.4 & 32.8 & 10.2 & 31.1 & 23.3 & 26.1 & 13.7 \\
Candidate verification & 55.0 & 51.7 & 36.7 & 37.4 & 53.3 & 53.3 & 31.1 & 36.9 \\
ReListen & \textbf{78.3} & \textbf{75.6} & \textbf{50.6} & \textbf{60.7} & \textbf{75.6} & \textbf{72.8} & \textbf{45.6} & \textbf{56.1} \\
\midrule
\grouprow{9}{Using the Q1 options}
Option rule, no listening to locate & 47.2 & 72.2 & 47.8 & 35.7 & 43.9 & 71.1 & 41.7 & 33.9 \\
Reader's Q1 choice & 18.3 & 58.9 & 42.2 & 12.0 & 27.2 & 61.7 & 41.1 & 18.7 \\
ReListen, chain-consistent & \textbf{78.3} & \textbf{81.7} & \textbf{53.9} & \textbf{62.8} & \textbf{75.6} & \textbf{80.6} & \textbf{47.8} & \textbf{59.3} \\
\midrule
\grouprow{9}{Reference}
Oracle time, smaller count & 67.2 & 88.9 & 56.1 & 52.2 & 63.3 & 87.8 & 51.1 & 48.9 \\
\bottomrule
\end{tabular}
\caption{\textbf{Occurrence chains on both halves.}
We report accuracy (\%) on each question and the chain score on 180 chains per half.
Every audio selection totals 60 seconds, and the transcript is from Whisper.
The chain-consistent row answers Q2 and Q3 at the time that ReListen chose for Q1.}
\label{tab:mainfull}
\end{table*}

\begin{table}[h]
\centering
\small
\setlength{\tabcolsep}{3pt}
\begin{tabular}{lccccc}
\toprule
 & Ins. & Noise & Sil. & Speed & Vol. \\
 & (60) & (31) & (29) & (26) & (34) \\
\midrule
\grouprow{6}{Time accuracy}
Reader evidence only & 61.7 & 90.3 & 82.8 & 3.8 & 2.9 \\
Detector evidence only & 35.0 & 87.1 & 82.8 & \textbf{88.5} & \textbf{73.5} \\
ReListen & \textbf{71.7} & \textbf{93.5} & \textbf{89.7} & \textbf{88.5} & \textbf{73.5} \\
\midrule
\grouprow{6}{Q2 accuracy}
Candidate verification & 56.7 & 77.4 & 51.7 & 38.5 & 29.4 \\
ReListen & \textbf{73.3} & \textbf{80.6} & \textbf{69.0} & \textbf{80.8} & \textbf{76.5} \\
Oracle time (reference) & 95.0 & 87.1 & 72.4 & 88.5 & 94.1 \\
\bottomrule
\end{tabular}
\caption{\textbf{Occurrence chains by edit type.}
We report time accuracy and Q2 accuracy (\%) on the test half.
Parentheses give the number of chains of each edit type.}
\label{tab:bytype}
\end{table}

Table~\ref{tab:mainfull} gives the chain results of Section~\ref{sec:chain-results} on both halves.
Table~\ref{tab:bytype} separates the test results by edit type, and the development half shows the same pattern.
In the first group of Table~\ref{tab:mainfull} the reader answers Q1 itself, except for candidate verification and ReListen, which answer it from their enumeration.
\emph{Reader's Q1 choice} hears again at the time of the option that the reader selected for Q1.
Time accuracy is the fraction of chains for which the selected time lies within 30 seconds of the midpoint of the $k$-th edited region.
On the chains ReListen exceeds the option rule by 25 points.
On all 180 test chains, the full recording reaches a chain score of 16.5, and 17.2 with time marks, against 60.7 for ReListen.

\paragraph{Remaining Errors.}
On the test half, 34 of 180 chains receive a wrong time.
Of these, 17 have inserted sounds, 9 volume changes, 3 each silence masks and speed changes, and 2 noise masks.
In 19 of them ReListen selects a false alarm, in 13 it selects a different true occurrence, and in 2 no candidate lies at the target.
The dominant failure is thus an ordinal error among detected candidates.
Q3 reaches 56.1\% at the oracle time.
Hence, a large share of the remaining chain errors comes from reasoning about content beyond the window, the same limit that the native questions show.

\section{Ablation of the Occurrence Path}
\label{app:ablation}

\begin{table}[h]
\centering
\small
\setlength{\tabcolsep}{2.5pt}
\begin{tabular}{lccccc}
\toprule
 & & \multicolumn{2}{c}{Test} & \multicolumn{2}{c}{Dev} \\
\cmidrule(lr){3-4}\cmidrule(lr){5-6}
Variant & Windows & Time & Q1 & Time & Q1 \\
\midrule
\grouprow{6}{Removing one component}
Reader evidence only & 289 & 50.6 & 51.1 & 54.4 & 53.3 \\
Detector evidence only & 289 & 66.7 & 71.1 & 72.2 & 71.1 \\
No calibration labels & 289 & 60.0 & 58.3 & 59.4 & 59.4 \\
Hard ordinal rule & 289 & 80.0 & 72.8 & 79.4 & 71.7 \\
Single detector seed & 289 & 75.7 & 71.9 & 81.3 & 72.2 \\
\midrule
\grouprow{6}{Scoring fewer windows}
20 s hop & 145 & 81.1 & 77.8 & 80.6 & 75.6 \\
Retriever-pruned scan & 92 & 79.4 & 76.1 & 81.1 & 76.1 \\
Retriever-pruned scan & 48 & 76.7 & 73.3 & 78.9 & 75.6 \\
Retriever-pruned scan & 16 & 74.4 & 70.6 & 77.2 & 73.3 \\
\quad reader only, hard rule & 16 & 45.0 & 43.3 & 49.4 & 48.9 \\
\midrule
ReListen & 289 & 81.1 & 78.3 & 80.6 & 75.6 \\
\bottomrule
\end{tabular}
\caption{\textbf{Ablation of the occurrence path.}
We report time accuracy and Q1 accuracy (\%).
Windows is the mean number of windows that the reader scores per test recording.
The detector-only row falls back to the reader's scan when the detectors propose no region.
The single-seed row averages three seeds.}
\label{tab:ablation}
\end{table}

Table~\ref{tab:ablation} removes one element of the occurrence path at a time.
The acoustic detectors matter most: without them time accuracy falls by 31 points on the test half.
By comparison, without the reader scan it falls by 14 points.

\paragraph{Calibration and Its Labels.}
Calibration is the third largest factor.
A fusion without benchmark annotation takes the noisy-OR of the reader's raw yes-probability and the detector probability.
It reaches 60.0\% time accuracy, 21 points below the calibrated system.
However, the calibrators need few annotated recordings.
Fitted on 12 development recordings drawn at random, they reach 72.8\,$\pm$\,3.2\% time accuracy on the test half over 20 draws, and 78.5\,$\pm$\,1.6\% on 96 recordings.
Fitted on the test half and applied to the development half, they yield 81.7\%, against 80.6\% with cross-fitting.

\paragraph{The Ordinal Posterior.}
Replacing the posterior by the hard rule leaves time accuracy within one point but lowers Q1 by 4 to 6 points on each half.
Pooled over both halves the posterior wins Q1 on 38 chains and loses on 21, a significant difference.
This gain comes from scoring each Q1 option by its count and its location together.
The posterior's most probable count alone is about as accurate as the count of the hard rule, 66.1\% against 68.3\% on the test half.

\paragraph{Ensemble and Cost.}
A single seed yields 81.3\% time accuracy on development recordings and 75.7\% on the test half.
Doubling the hop to 20 seconds halves the reader's work and changes Q1 on five test chains, three for the worse and two for the better.
Restricting the scan to the windows ranked highest by the learned retriever trades accuracy for cost.
With 16 windows per recording, time accuracy on the test half is 74.4\% with the detectors.
It is 45.0\% in the configuration that corresponds to candidate verification.

\section{Occurrence Path beyond Planted Edits}
\label{app:natural}

\paragraph{Unedited Recordings.}
The detectors could in principle respond to natural events in the recordings and not to the edits.
To check this, we ran the detector branch of the configuration on the unedited version of every recording with the event of its chain.
The expected number of occurrences is 0.52 per development recording and 0.56 per test recording on the unedited versions.
On the edited versions it is 1.76 and 1.80, where the true mean is 1.94.
Similarly, a candidate with a calibrated probability above one half exists in 17\% and 15\% of the unedited recordings and in 80\% and 83\% of the edited ones.

\paragraph{A Natural Event.}
The AMI meeting corpus \citep{carletta2006ami} annotates coughs.
We built 119 questions over 24.5 hours about the $k$-th cough for the 49 meetings with one to four occurrences.
Coughs closer than 30 seconds count as one occurrence.
We then ran the reader's presence scan with the phrase ``a person coughing'', without detectors.
The window-level AUC of the scan is 0.88.
With the hard rule on decoded answers, the selected time is within 30 seconds of the $k$-th cough for 37.0\% of the questions, against 3.8\% for a random time.
The count is exact for 26.5\% of the meetings.
By comparison, the calibrators fitted on the benchmark's inserted sounds and the posterior yield 36.1\% and 26.5\%.
Calibrators cross-fitted on the meetings yield 37.8\% and 24.5\%.
The scan thus finds a natural sound far above chance, whereas calibration and the posterior add nothing measurable at this size.

\end{document}